\documentclass[letterpaper,journal]{IEEEtran}
\usepackage{amsmath,amsfonts}
\usepackage{algorithmic}
\usepackage{algorithm}
\usepackage{array}
\usepackage[caption=false,font=normalsize,labelfont=sf,textfont=sf]{subfig}
\usepackage{textcomp}
\usepackage{stfloats}
\usepackage{url}
\usepackage{graphicx}
\usepackage[nocompress]{cite}
\usepackage{booktabs}
\usepackage[hidelinks]{hyperref}
\hypersetup{
  pdftitle={WinDEX: Metric 3D Hand Motion Reconstruction in the Wild for Dexterous Teleoperation and Policy Learning},
  pdfauthor={Hongyu Ma, Hairong Qu, Shiqi Zhao, Yongsong Yang, Kai Shi, Jiahao Ai, Xin You, Peng Yin},
  pdfsubject={Metric world-frame 3D hand reconstruction from egocentric stereo video},
  pdfkeywords={3D hand pose estimation, world frame, egocentric stereo vision, learning from demonstration, teleoperation}
}

\newcommand{\degs}{\ensuremath{^{\circ}}}

\begin{document}
\bstctlcite{IEEE:BSTcontrol}

\title{WinDEX: Metric 3D Hand Motion Reconstruction in the Wild\\
for Dexterous Teleoperation and Policy Learning}

\author{\begin{minipage}{\textwidth}
\centering\normalfont\fontsize{11}{13}\selectfont
Hongyu~Ma$^{1}$, Hairong~Qu$^{1}$, Shiqi~Zhao$^{1}$, Yongsong~Yang$^{1}$,\\[2pt]
Kai~Shi$^{1}$, Jiahao~Ai$^{1}$, Xin~You$^{2}$, and Peng~Yin$^{1}$\par
\vspace{0.25em}
{\normalsize $^{1}$City University of Hong Kong \qquad $^{2}$CyberOrigin}
\end{minipage}}

\markboth{WinDEX: Metric 3D Hand Motion Reconstruction in the Wild}{Ma et al.: WinDEX}

\renewcommand{\IEEEtitletopspaceextra}{-10pt}
\IEEEaftertitletext{\vspace{-1.5\baselineskip}}
\maketitle

\begin{abstract}
Human manipulation data are widely used to train robots for human-like dexterity.
However, low-cost, large-scale collection in the wild remains challenging.
Existing methods remain limited in geometrically grounded metric-scale reconstruction, real-world generalization and inference speed.
We present WinDEX, integrating (1) geometric--temporal optimization for metric 3D hand annotation from in-the-wild egocentric stereo videos, combined with visual--inertial SLAM for world-frame reconstruction;
(2) WinDEX-Net, an end-to-end, fisheye-native network combining calibrated stereo, local detail and temporal context, with a data--model flywheel of model-assisted annotation and retraining; (3) distilled models for offline reconstruction and causal streaming, supporting reinforcement learning and teleoperation;
and (4) low-cost CyberCap stereo--inertial capture, forming the WinDEX-Hand dataset with 1{,}000 hours of training annotations and held-out human-annotated in-the-wild 2D and independent optical motion-capture evaluation sets.
Offline reconstruction runs at 43.61 FPS and causal streaming at 171 FPS.
On held-out data, offline reconstruction achieves state-of-the-art accuracy: 14.5 pixels in-the-wild reprojection error and, with visual--inertial odometry, 37.29 mm world-frame joint error.
We achieve $7.6\times$ and $3.6\times$ the task success rates of reinforcement learning policies trained with WiLoR and HaWoR motion references, respectively, and demonstrate real-time arm--hand teleoperation.
Project website: \url{https://fruitrabbit.github.io/WinDEX/}.
\end{abstract}


\section{Introduction}\label{sec:intro}

\begin{figure}[!t]
\centering
\includegraphics[width=\columnwidth]{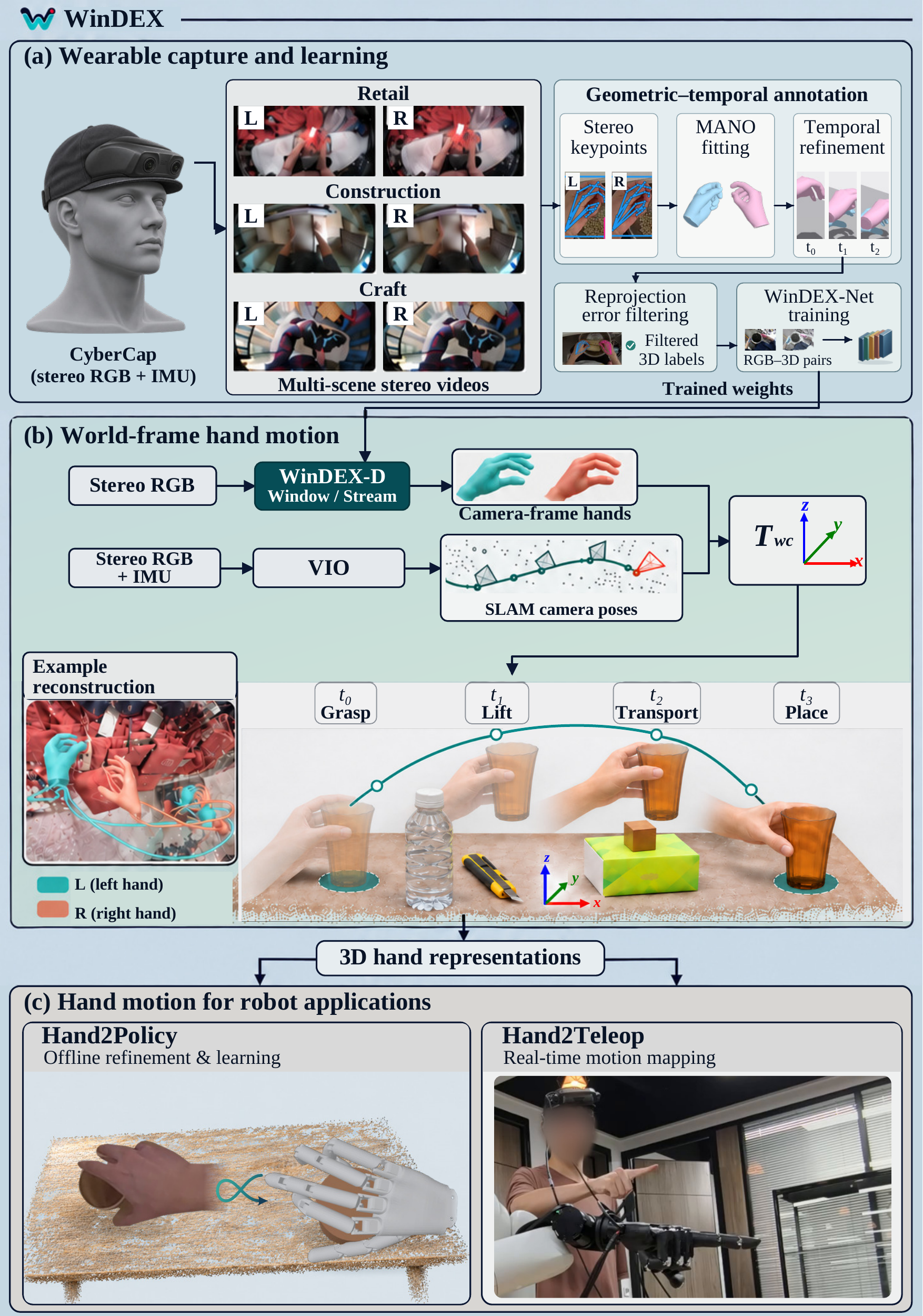}
\caption{WinDEX overview. (a) CyberCap records stereo and inertial data for geometric--temporal annotation and WinDEX-Net training. (b) WinDEX-D hand estimates are combined with VIO camera poses to produce world-frame 3D hand representations. (c) Hand2Policy uses offline refinement for policy learning, and Hand2Teleop maps motion to live robot control.}
\label{fig:teaser}
\end{figure}

\IEEEPARstart{R}{oboticists} have long sought to guide dexterous robotic
manipulation using human motion trajectories. Hand poses and trajectories guide
teleoperation
\cite{ref:handmap,ref:anyteleop,ref:televison,ref:dexcap} and provide
demonstrations, pose priors, and tracking references for reinforcement learning (RL)
\cite{ref:dexmv,ref:dexvip,ref:maniptrans}. Hand trajectories also provide
action supervision for video-based policies and vision--language--action
models \cite{ref:egomimic,ref:egovla,ref:beingh0}, condition world models that
predict manipulation outcomes \cite{ref:dwm}, and supply geometric
information for grasp synthesis \cite{ref:graspreview,ref:manipnet}.
These applications share a need for reliable 3D hand motion in relation to
the surrounding scene. We adopt metric 3D hand poses and their trajectories
in a gravity-aligned world frame as our 3D hand representations, describing
wrist motion and finger articulation independently of the moving camera
(Fig.~\ref{fig:teaser}). Recovering this representation across everyday
environments at scale is our central problem.

Existing acquisition routes include optical motion capture, calibrated
multi-camera rigs, sensor gloves and wearable trackers, and video recording.
Optical systems and multi-camera datasets such as ARCTIC,
DexYCB, HOT3D and AssemblyHands provide detailed reference measurements
\cite{ref:arctic,ref:dexycb,ref:hot3d,ref:assemblyhands}, but depend on
instrumented environments. Gloves and exoskeletons support portable capture
\cite{ref:dexcap,ref:dexumi} while retaining hardware on the hands.

Monocular RGB cameras permit egocentric or exocentric recording without
hand instrumentation. More broadly, video collections such as Ego4D and
Ego-Exo4D cover diverse first- and third-person activity
\cite{ref:ego4d,ref:egoexo4d}, but a single view leaves depth and absolute
scale ambiguous, and external cameras limit recording to their fields of
view. Wearable recording extends coverage
\cite{ref:nymeria,ref:egodex}. In particular, head-worn stereo can follow the
wearer without placing equipment on the hands, while a calibrated baseline
provides a geometric constraint on metric depth. Project Aria and Quest
tracking systems demonstrate the potential of wearable multi-view sensing
\cite{ref:aria,ref:megatrack,ref:umetrack}. Scalable in-the-wild capture still
requires reliable annotation under occlusion, peripheral visibility and
simultaneous camera and hand motion.

To recover metric hand trajectories from these recordings, both model
architecture and training supervision must accommodate wearable capture.
Monocular image and video estimators, including HaMeR, WiLoR, HaWoR and
Dyn-HaMR \cite{ref:hamer,ref:wilor,ref:hawor,ref:dynhamr}, use appearance
and motion priors to resolve single-view ambiguity. POEM and its successor
exploit calibrated multi-view geometry \cite{ref:poem,ref:poemv2};
ViDiHand integrates localization and reconstruction without an external
detector \cite{ref:vidihand}. For compact fisheye stereo, an integrated
architecture must additionally use the known baseline and viewing rays,
preserve local finger detail, and distinguish hand motion from camera
motion. On the data side, laboratory capture limits environmental coverage,
and tracker-generated labels require independent accuracy evaluation
\cite{ref:hot3d,ref:egodex}. Dense, reliable 3D supervision across diverse
natural recordings remains costly. These needs motivate joint localization
and reconstruction with stereo and temporal context, coupled to a scalable
pipeline for producing and checking training annotations.

\begin{figure*}[!t]
\centering
\includegraphics[width=\textwidth]{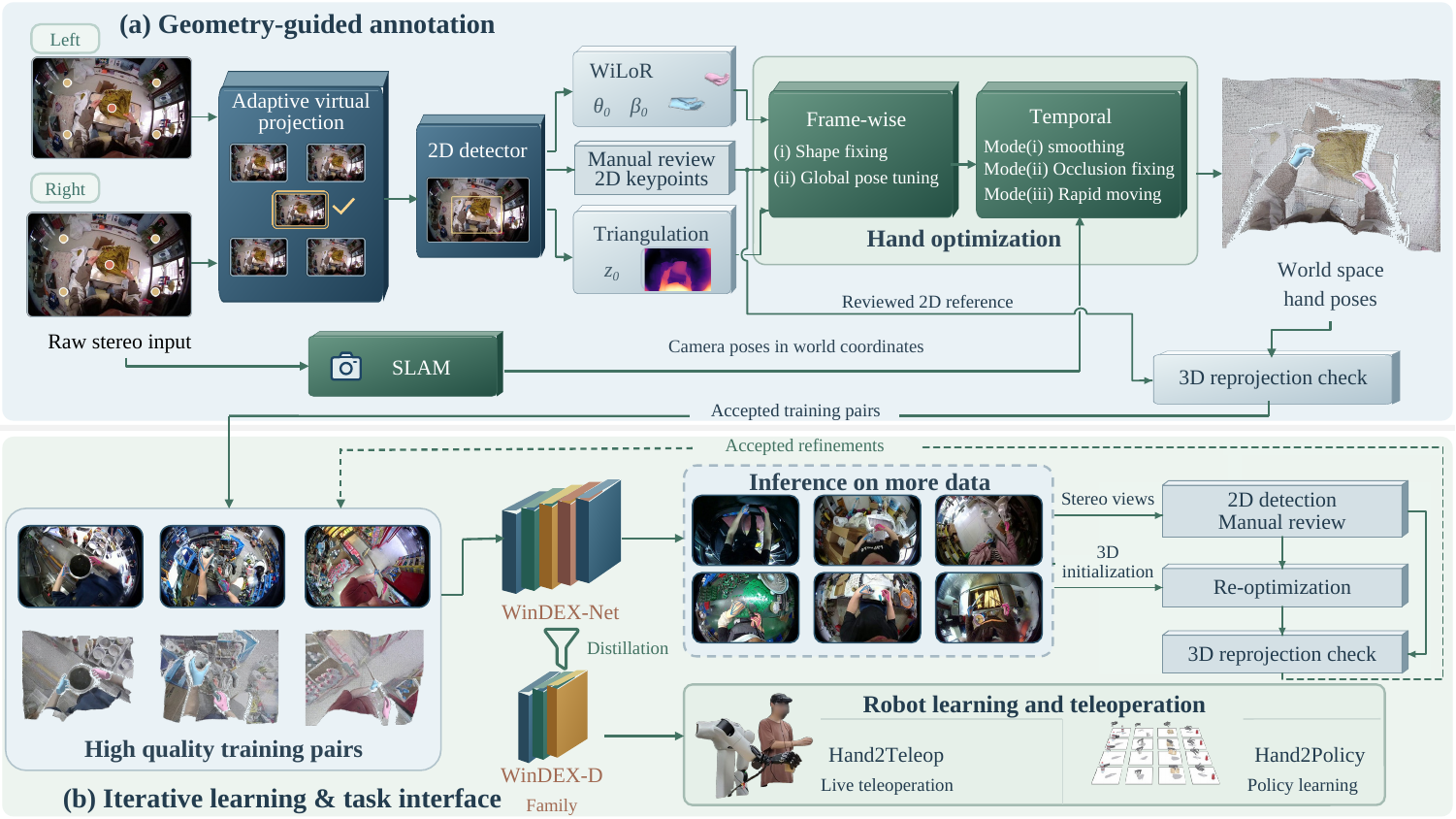}
\caption{System overview. (a) Geometry-guided annotation: manually reviewed 2D keypoints guide hand fitting; reprojection-error filtering selects 3D annotations. (b) Iterative learning \& task interface: the same review and filtering protocol supports the data--model flywheel, and distilled models serve Hand2Policy and Hand2Teleop.}
\label{fig:system}
\end{figure*}

We present WinDEX, a system integrating wearable capture,
geometric--temporal annotation, learned reconstruction and robot interfaces
(Fig.~\ref{fig:system}). Low-cost CyberCap collects fisheye stereo RGB and
inertial data without hand instrumentation. To obtain metric supervision, geometric--temporal
optimization fits hands to manually reviewed 2D keypoints, filters 3D labels
by reprojection error, and uses SLAM camera trajectories to recover
world-frame motion.

To address depth ambiguity, preserve finger detail and account for camera
motion, WinDEX-Net combines calibrated stereo, local crops and
odometry-conditioned temporal features for end-to-end, fisheye-native
localization and reconstruction. Predictions accelerate annotation, and
retraining closes the data--model flywheel. WinDEX-Hand separates generated 3D training labels
from evaluation references: human-annotated in-the-wild 2D references
and independent optical 3D ground truth (Section~\ref{sec:dataset}).

To balance accuracy and response time, distillation yields offline
WinDEX-Window and causal, real-time WinDEX-Stream. Hand2Policy uses
pose-tracking rewards for reinforcement learning; Hand2Teleop coordinates
arm--hand teleoperation using only CyberCap stereo and IMU for operator tracking.

Against human-annotated in-the-wild 2D references and optical 3D ground
truth from held-out recordings,
WinDEX-Window achieves state-of-the-art reconstruction accuracy and temporal
smoothness. Under matched
training settings and budgets, Hand2Policy reaches 61\% mean simulated
grasp--lift--place success versus 8\% for WiLoR and 17\% for HaWoR, with
policy transfer to physical robots.

The main contributions of this work are as follows:
\begin{enumerate}
\item To obtain metric 3D hand annotations from in-the-wild egocentric stereo
videos, we combine geometric--temporal optimization with visual--inertial
SLAM to recover hand articulation and metrically scaled trajectories in a
consistent world frame.
\item To accelerate reconstruction, we introduce WinDEX-Net, an end-to-end,
fisheye-native network combining calibrated stereo geometry, local hand
detail and temporal context. Model-initialized optimization and retraining
form a data--model flywheel that expands supervision.
\item To support offline learning and live control, we distill WinDEX-Net
into WinDEX-Window and WinDEX-Stream. Hand2Policy and Hand2Teleop further connect
the models to reinforcement learning with policy transfer and real-time
arm--hand teleoperation, respectively, with validation on physical hardware.
\item To make collection accessible at scale, we use the low-cost CyberCap
stereo--inertial device for everyday manipulation and construct WinDEX-Hand
from the annotated recordings, with held-out human-annotated in-the-wild 2D
references and independent optical motion-capture ground truth.
\end{enumerate}

Section~\ref{sec:related} reviews related work, and
Section~\ref{sec:overview} formulates the system. Sections~\ref{sec:capture}
and~\ref{sec:flywheel} describe metric annotation and model-assisted corpus
construction. Section~\ref{sec:deliver} presents the deployment models and
robot interfaces; Section~\ref{sec:downstream} reports experiments, followed
by limitations in Section~\ref{sec:limits}.

\section{Related Work}\label{sec:related}

\begin{table*}[!t]
\caption{Comparison of egocentric datasets and hand-motion supervision.
The WinDEX-Hand row describes its in-the-wild subset; validation uses held-out
human-annotated in-the-wild 2D references for image consistency and independent
optical motion-capture ground truth for 3D accuracy.}
\label{tab:positioning}
\centering
\resizebox{\textwidth}{!}{%
\begin{tabular}{llcccc}
\hline
Corpus & Setting & Metric scale from & Hand output & World frame & Accuracy validated\\
\hline
Ego4D \cite{ref:ego4d} & in the wild & --- & 2D keypoints & no & not reported\\
Ego-Exo4D \cite{ref:egoexo4d} & in the wild & calibrated multi-view rig & 3D joints & yes & manual triangulation checks\\
ARCTIC \cite{ref:arctic} & laboratory & multi-camera rig & MANO & rig frame & mocap ground truth\\
HOT3D \cite{ref:hot3d} & laboratory & instrumented rig & MANO + joints & rig frame & mocap ground truth\\
Nymeria \cite{ref:nymeria} & in the wild & body mocap suit & body, no fingers & yes & no\\
EgoDex \cite{ref:egodex} & indoor & Apple Vision Pro (ARKit) & joint transforms & yes & not reported\\
EgoScale (wild subset) \cite{ref:egoscale} & in the wild & not specified & 21 joint transforms & yes, estimated poses & not reported\\
EgoVerse-A (public subset) \cite{ref:egoverse} & in the wild & Project Aria Gen 1 (MPS) & 21 keypoints & via MPS SLAM poses & not reported\\
\hline
WinDEX-Hand (ours) & in the wild & calibrated RGB stereo baseline & MANO + metric wrist & yes, stereo VIO & \begin{tabular}[c]{@{}c@{}}optical mocap (3D)\\human-annotated wild 2D\end{tabular}\\
\hline
\end{tabular}}
\end{table*}

\subsection{Human Hand Motion for Robot Control and Learning}\label{subsec:consumers}

Human hand motion supports teleoperation, policy learning, and action
representations for vision--language--action (VLA) and world models.
AnyTeleop \cite{ref:anyteleop} converts visual hand estimates into robot
commands through learning-free retargeting and collision-aware motion
generation. DexCap \cite{ref:dexcap} matches human and robot fingertips
through inverse kinematics and learns point-cloud diffusion policies from
wearable demonstrations. For policy learning, DexMV \cite{ref:dexmv}
reconstructs hand--object poses from paired RGB-D views in a configured
recording volume, translates them into robot states and actions, and
trains policies in corresponding simulated tasks. ManipTrans
\cite{ref:maniptrans} pretrains a generalist trajectory imitator and learns
residual corrections for object interaction, assuming supplied hand--object
trajectories. EgoVLA \cite{ref:egovla} and Being-H0 \cite{ref:beingh0} use
hand trajectories as action supervision for VLA pretraining from human
videos. Dexterous World Models \cite{ref:dwm} condition visual-dynamics
prediction on hand motion. These uses depend on reliable motion
representations. WinDEX addresses this shared need by recovering metric,
world-frame hand motion from everyday scenes.

\subsection{Hand Motion Capture and Egocentric Datasets}\label{subsec:scale}

Hand-motion datasets differ in capture hardware, image modality and label
reliability. DexCap combines
electromagnetic gloves with SLAM cameras \cite{ref:dexcap}; DexUMI uses
joint encoders in a robot-specific exoskeleton and ARKit wrist tracking
\cite{ref:dexumi}. Both retain hardware on the hands.

Headset datasets also differ from our RGB stereo setting. UmeTrack uses four monochrome fisheye
cameras and marker-based optical ground truth \cite{ref:umetrack}.
HOT3D records Aria and Quest 3 streams with optical ground truth in an
instrumented laboratory; their stereo tracking streams are monochrome,
with an additional RGB view on Aria \cite{ref:hot3d}.
EgoDex scales recording through Apple Vision Pro's ARKit tracker
\cite{ref:egodex}, whose predicted labels can degrade under occlusion and
rapid motion. It supplies joint confidence but does not report independent
motion-capture validation or a reprojection-error acceptance threshold.
We evaluate on held-out WinDEX-Hand recordings, using independent optical
ground truth for 3D accuracy and human-annotated 2D references for
in-the-wild image consistency. Table~\ref{tab:positioning} compares the
capture and supervision choices.

\subsection{Image-Based 3D Hand Reconstruction}\label{subsec:estimators}

Many image-based pipelines first localize the hand and then estimate its
pose from a cropped image \cite{ref:megatrack,ref:umetrack,ref:assemblyhands}.
Monocular image and video methods
\cite{ref:hamer,ref:wilor,ref:hawor,ref:dynhamr} recover detailed hand pose
and shape using learned appearance and motion priors. Metric depth remains
dependent on additional assumptions because a single view does not directly
determine absolute scale. POEM and POEM-v2 \cite{ref:poem,ref:poemv2}
use calibrated observations to constrain 3D reconstruction, whereas
HGGT \cite{ref:hggt} jointly estimates hand meshes and camera poses from
uncalibrated images. Applying multi-view reconstruction to a head-worn stereo pair
requires accounting for a short baseline, fisheye distortion, coupled
camera and hand motion, and occlusion.

Whole-image estimation \cite{ref:hplusO} and recent video-based methods
\cite{ref:vidihand,ref:dreamhand} explore reconstruction without a separate
hand-detection stage. WinDEX-Net builds on this direction for native fisheye
stereo input. It combines calibration-aware features with temporal context
to estimate hand articulation and metric wrist position. Hand localization
and iterative crop refinement are integrated within the network, avoiding
dependence on a separately trained detector. The architecture is described
in Section~\ref{sec:model}; the geometric basis for metric reconstruction
is given in Section~\ref{sec:capture}.

\subsection{Model-Assisted Annotation and Iterative Learning}\label{subsec:engines}

Model-assisted data collection and self-training couple annotation with
repeated model updates \cite{ref:sam,ref:noisystudent}. These strategies
can extend supervision to larger datasets, while introducing the risk that
prediction errors propagate across training rounds. WinDEX uses model
predictions to initialize annotation, then refines them against stereo
observations before adding accepted results to the training set
(Section~\ref{sec:coevolve}). Geometric optimization and filtering by
reprojection error constrain labels using the recorded image evidence. A separate
motion-capture set, excluded from training, provides an independent measure
of model accuracy throughout the iterations. This combination links
annotation scale to geometric consistency and external evaluation.

\section{System Overview and Problem Formulation}\label{sec:overview}

WinDEX connects egocentric recordings of everyday manipulation to robot
control and learning through metric, world-frame hand motion. Extending
egocentric corpora \cite{ref:ego4d,ref:egoexo4d,ref:egodex} with this
representation requires both geometric reconstruction and annotation at
scale. Calibrated stereo supplies depth constraints, and inertial
measurements support camera tracking. An offline optimizer uses these
observations to generate supervision for WinDEX-Net; model predictions then
initialize further annotation. Distillation yields WinDEX-Window for
offline reconstruction in Hand2Policy and WinDEX-Stream for causal
inference in Hand2Teleop.\label{subsec:overview}

\subsection{Problem Formulation}

Let a recording contain $T$ synchronized stereo fisheye frame pairs
$\mathcal{I}=\{I^{\,l}_t,I^{\,r}_t\}_{t=1}^{T}$, known equidistant
intrinsics and inter-eye extrinsics $(\mathtt{R}_{lr},\mathbf{t}_{lr})$
with $\|\mathbf{t}_{lr}\|=b$, and the rig's inertial stream $\mathcal{A}$.
Capture requires no external cameras, markers, gloves or prior scene
or task information. The hand state is
$\mathbf{x}_t=(\theta_t,\mathbf{w}_t)$, where $\theta_t$ contains MANO
\cite{ref:mano} pose parameters and $\mathbf{w}_t$ is the wrist translation
in the reference camera frame, in meters. A single shape vector $\beta$ is
shared across each hand's recording. For both hands, the outputs are the
hand-state trajectory and world-frame camera poses,
\begin{equation}
\mathcal{X}=\big\{\theta_t,\ \beta,\ \mathbf{w}_t\big\}_{t=1}^{T},\qquad
\mathcal{T}=\big\{T^{\,w}_{c,t}\big\}_{t=1}^{T},
\label{eq:state}
\end{equation}
whose composition places the camera-frame hand estimates in a
gravity-aligned world frame. The resulting hand poses and trajectories
constitute the 3D hand representations used by Hand2Policy and Hand2Teleop.

We organize reconstruction as a factorized maximum-a-posteriori (MAP)
approximation. Detected 2D keypoints $\mathbf{u}^{l}_t,\mathbf{u}^{r}_t$
provide the image observations, and camera motion is estimated separately
from stereo and inertial measurements:
\begin{equation}
\begin{split}
\tilde p(\mathcal{X},\mathcal{T}\mid\mathcal{I},\mathcal{A})\ \propto\
&\ p(\mathcal{T}\mid\mathcal{I},\mathcal{A})\,p_{\beta}(\beta)\\[-2pt]
&\times\prod\nolimits_{t=1}^{T} p_{\mathrm{obs}}\big(\mathbf{u}^{l}_t,\mathbf{u}^{r}_t
 \mid\theta_t,\beta,\mathbf{w}_t\big)\,p_{\mathrm{pose}}(\theta_t)\\[-2pt]
&\times\prod\nolimits_{t=3}^{T} p_{\mathrm{dyn}}(\mathbf{x}_t\mid\mathbf{x}_{t-1},\mathbf{x}_{t-2}) .
\end{split}
\label{eq:map}
\end{equation}
Here $p_{\mathrm{obs}}$ describes binocular observations,
$p_{\mathrm{pose}}$ and $p_{\beta}$ encode hand priors, and
$p_{\mathrm{dyn}}$ models temporal regularity.

For conditionally independent observations in the two eyes,
\begin{equation}
\begin{split}
p_{\mathrm{obs}}\big(\mathbf{u}^{l}_t,\mathbf{u}^{r}_t\mid\theta_t,\beta,\mathbf{w}_t\big)
=\prod_{e\in\{l,r\}}
p_{\mathrm{obs}}\big(\mathbf{u}^{e}_t\mid\pi^{e}_{\mathrm{KB}}(\mathbf{P}_t)\big),\\
\mathbf{P}_t=\mathcal{J}(\theta_t,\beta)+\mathbf{w}_t .
\end{split}
\label{eq:stereo}
\end{equation}
Here $\mathcal{J}$ returns wrist-centered MANO joints, and $\pi^{e}_{\mathrm{KB}}$ is the
calibrated fisheye projection; the right-eye projection first transforms
all joints by $\mathtt{R}_{lr}\mathbf{P}_t+\mathbf{t}_{lr}$. The known
baseline constrains metric placement.

\subsection{System Design}

Reliably modeling every factor in (\ref{eq:map}) across in-the-wild
conditions is difficult. We therefore learn reconstruction from supervision
obtained through tractable geometric and hand-prior constraints. The
workflow proceeds through data collection, annotation, learned inference
and robot deployment.

\begin{enumerate}
\item Low-Cost Capture and WinDEX-Hand.
To make collection accessible at scale, we use low-cost CyberCap to record
synchronized stereo images and inertial measurements during everyday
manipulation. WinDEX-Hand organizes recordings for annotation and model
training, with held-out human-annotated in-the-wild 2D references and independent optical
motion-capture ground truth for evaluation
(Sections~\ref{subsec:collection} and~\ref{sec:dataset}).

\item Geometric--Temporal Annotation.
To obtain metric supervision, we fit hand motion to manually reviewed 2D
keypoints under the observation, hand-prior and temporal factors in
(\ref{eq:map}). Reprojection-error filtering selects 3D annotations, and
visual--inertial SLAM camera trajectories place hand motion in the world
frame (Section~\ref{sec:capture}).

\item WinDEX-Net and the Data--Model Flywheel.
To accelerate reconstruction, WinDEX-Net combines calibrated stereo, local
hand detail and odometry-conditioned temporal context in an end-to-end,
fisheye-native model. It approximates the hand-state MAP solution in
(\ref{eq:map}), with SLAM camera poses held fixed. Predictions initialize
further optimization against manually reviewed 2D observations; refined 3D
labels passing the same reprojection-error threshold expand training,
closing the data--model flywheel (Section~\ref{sec:flywheel}).

\item Distillation and Robot Deployment.
To balance accuracy, computational cost and response time, we distill
WinDEX-Net into WinDEX-D. WinDEX-Window uses bidirectional context for
accurate offline reconstruction; WinDEX-Stream retains causal state for
real-time inference without future frames. Combined with SLAM camera
poses, the estimates support Hand2Policy for trajectory-guided reinforcement
learning and policy transfer to physical robots, and Hand2Teleop for live
arm--hand teleoperation (Section~\ref{sec:deliver}).
\end{enumerate}

\section{Data Collection and Geometric--Temporal Annotation}\label{sec:capture}

We recover hand articulation and metric placement by combining stereo
evidence, temporal hand priors and visual--inertial camera poses. The
following collection and annotation stages implement these factors and
produce filtered training supervision.

\subsection{Recording Configuration and Data Collection}\label{subsec:collection}

The binocular observation term in (\ref{eq:stereo}) requires synchronized,
calibrated views of the same hand; the camera-pose factor in
(\ref{eq:map}) requires aligned visual and inertial measurements.
CyberCap supplies both. Its cameras share a calibrated
60.7\,mm baseline and capture hardware-synchronized RGB fisheye images
at $1920\times1200$ pixels per eye and 30\,fps. A rigidly mounted IMU
records at 300\,Hz on the same clock, with measurements embedded in the
video stream. Calibrated fisheye intrinsics \cite{ref:kb}, stereo
extrinsics and camera--IMU extrinsics connect these observations
geometrically.

We adapt occupational and work-activity information from O*NET
\cite{ref:onet} to organize recording activities into collection categories.
WinDEX-Hand separates training recordings from human-annotated 2D-only
in-the-wild evaluation and independently captured optical motion-capture
ground truth; Section~\ref{sec:dataset} details their composition and splits.

\subsection{World-Frame Odometry}\label{subsec:vio}

Hand reconstruction alone describes motion relative to a moving camera.
We estimate the camera-pose factor $p(\mathcal{T}\mid\mathcal{I},\mathcal{A})$
in (\ref{eq:map}) with cuVSLAM \cite{ref:cuvslam}, combining
calibrated stereo observations with inertial measurements to constrain
camera motion during occlusion and rapid head turns. This follows
visual--inertial estimation \cite{ref:vinsmono,ref:orbslam3} using native
fisheye images and the camera and IMU calibration. Poses take the first
recorded frame as the world origin and carry covariance and confidence
estimates; initial poses remain low-confidence until gravity estimation
settles.

Composing camera and hand poses makes odometry error part of the
world-frame reconstruction error. Against independent optical motion
capture, the camera trajectory has a root mean square absolute trajectory
error of 0.150\,m and a root mean square orientation error of
11.69$^\circ$.

\subsection{Annotation by Optimization}\label{sec:annotate}

The observation factor
$p_{\mathrm{obs}}(\mathbf{u}^{l}_t,\mathbf{u}^{r}_t\mid\theta_t,\beta,\mathbf{w}_t)$
in (\ref{eq:map}) makes binocular 2D keypoints the reference evidence for
fitting. Pose estimators supply these observations, and human review
selects reliable 2D keypoints. Fitted 3D annotations are then filtered by
their reprojection error against the accepted keypoints. Pose, shape and motion priors constrain
ambiguities in image evidence. The pipeline (Fig.~\ref{fig:system}(a))
therefore fits shared shape and per-frame pose and placement, then refines
motion across frames.

\subsubsection{Adaptive Virtual Projection}

The available hand detectors and pose estimators operate on pinhole
images. In our recordings, hands frequently occupy the fisheye periphery
(Fig.~\ref{fig:datastats}c). A single centered virtual view either omits
these regions or stretches them near its boundaries. We therefore render
five virtual pinhole views centered on the image center and four quadrant
centers. YOLO11-pose \cite{ref:yolo11} and ViTPose \cite{ref:vitpose}
estimate hand keypoints in these views, and the highest-confidence hand
detection selects the virtual view for each hand. This provides suitable
input to the estimators while retaining native fisheye coordinates for
the subsequent fitting objective.

Within the selected virtual view, we fuse the two estimators joint by
joint. Candidates with confidence
below 0.3 are discarded. If both remaining estimates lie within 10\% of
the hand-box side length, their coordinates are averaged with confidence
weights; otherwise, we retain the higher-confidence estimate, using the
hand-box score to resolve near-ties. Joints without a valid candidate are
excluded. The fused keypoints are mapped to native fisheye coordinates
and checked for stereo depth and hand-structure consistency before fitting.

\subsubsection{Initialization}

Initialization combines an appearance-based hand estimate with a
geometric depth estimate. WiLoR \cite{ref:wilor} predicts articulation
and shape $(\theta_0,\beta_0)$ from the selected view in each eye;
triangulated wrist disparity from a stereo matcher provides metric depth
$z_0$. The fused keypoints supply the observations in (\ref{eq:stereo}),
with residuals evaluated under the calibrated fisheye projection model.

\subsubsection{Stage One: Frame-Wise Optimization}

The first stage uses WiLoR's pose and shape estimates
$(\theta_0,\beta_0)$ and the triangulated depth $z_0$ as optimization
initial values, then fits MANO parameters and depth to the keypoints.
MANO defines the articulated geometry used to evaluate $p_{\mathrm{obs}}$, and DPoser-X
\cite{ref:dposerx} supplies the articulation prior $p_{\mathrm{pose}}$
in (\ref{eq:map}). A single shape vector $\beta$ is estimated for each
recording: allowing shape to vary frame by frame would let it absorb
depth errors and violate the shared-shape state in (\ref{eq:state}).
We estimate $\beta$ by
independently fitting the 64 highest-confidence frames, which need not
be consecutive, and averaging their shape estimates. With $\beta$
fixed, each frame is then optimized for articulation $\theta_t$, global
orientation and depth $z_t$.

\subsubsection{Stage Two: Temporal Optimization in Three Modes}

The motion factor $p_{\mathrm{dyn}}$ in (\ref{eq:map}) must suppress
fluctuations without erasing rapid hand motion. We use three modes---smooth
motion, occlusion and rapid motion---to adapt its strength relative to the
observation term in (\ref{eq:energy}). Smooth motion is the default.
Low 2D keypoint confidence, even in the better-observed eye, indicates
occlusion. When 2D confidence remains high but reprojection error is large,
we select rapid-motion mode: reliable image observations disagree with the
current temporally constrained estimate, suggesting that smoothing may be
restricting a fast pose change.

For smooth motion, temporal regularization and a 6\,Hz
low-pass filter reduce detector jitter. During occlusion, we hold
the last high-confidence pose until a new high-confidence frame becomes
available. This avoids unstable fits but does not recover unobserved
articulation. For rapid motion, weaker temporal weights allow the
2D observations to preserve fast changes and limit smoothing-induced lag.
The refined hand poses are composed with visual--inertial camera poses
(Section~\ref{subsec:vio}) to obtain world-frame trajectories.

\subsubsection{Energy}

The product in (\ref{eq:map}) jointly constrains one trajectory through
observations and hand priors. Taking negative logarithms yields additive
penalties, implemented using the geometry in (\ref{eq:stereo}) as the
weighted energy
\begin{align}
E(\mathcal{X}) =\ & \lambda_{\mathrm{2D}}
  \sum_{t}\sum_{e\in\{l,r\}}
  \rho_{\mathrm{GM}}\!\big(\pi_{\mathrm{KB}}^{e}(\mathcal{J}(\theta_t,\beta)+\mathbf{w}_t)
  - \mathbf{u}^{e}_{t}\big) \nonumber\\
 &+ \lambda_{\mathrm{st}} \sum_{t}
   \big\|\mathbf{w}_t-\hat{\mathbf{w}}_t\big\|^2
 \ +\ \lambda_{\mathrm{pose}} \sum_{t} E_{\mathrm{prior}}(\theta_t) \nonumber\\
 &+ \lambda_{\mathrm{dyn}} \sum_{t}
   \left(\big\|\Delta^2 \theta_t\big\|^2
   + \big\|\Delta^2 \mathbf{w}_t\big\|^2\right)
 \ +\ \lambda_{\beta}\|\beta\|^2 ,
\label{eq:energy}
\end{align}
where $\mathbf{u}^{e}_{t}$ denotes detected keypoints in eye $e$, $\rho_{\mathrm{GM}}$
is a Geman--McClure kernel, $\hat{\mathbf{w}}_t$ is the stereo-initialized
wrist, and $\Delta^2$ is the second temporal difference. Reprojection
represents $p_{\mathrm{obs}}$; $E_{\mathrm{prior}}$, $\|\beta\|^2$ and
the second differences encode $p_{\mathrm{pose}}$, $p_\beta$ and
$p_{\mathrm{dyn}}$, respectively. The shared shape and temporal terms
couple the frame-wise fits. The stereo wrist term supplies an additional
geometric anchor. Stage one sets
$\lambda_{\mathrm{dyn}}=0$; stage two adjusts
$\lambda_{\mathrm{dyn}}$ and $\lambda_{\mathrm{2D}}$ according to the
motion mode. The pose prior is trained outside this dataset and provides
a constraint independent of the flywheel's generated labels
(Section~\ref{sec:coevolve}).

\subsubsection{Filtering by Reprojection Error}

We filter fitted 3D hand states against the manually accepted 2D keypoints
in rectified pinhole views. Let $\mathcal{V}_t$ contain valid observations
$(e,j)$ across both eyes with 2D confidence at least 0.3.
Observations $\bar{\mathbf{u}}^{e}_{t,j}$
and fitted-hand projections use the same rectification and pixel scale;
$\pi^{e}_{\mathrm{rect}}$ includes the inter-eye and rectification transforms.
The reprojection error $\rho_t$ is their
mean Euclidean distance in pixels:
\begin{equation*}
\rho_t=\frac{1}{|\mathcal{V}_t|}
\sum_{(e,j)\in\mathcal{V}_t}
\big\|\pi^{e}_{\mathrm{rect}}(\mathbf{P}_{t,j})-\bar{\mathbf{u}}^{e}_{t,j}\big\|_2 .
\end{equation*}
We retain a hand-frame when $\rho_t\le\tau$, with a fixed acceptance
threshold $\tau=15$ pixels. Hand-frames without valid observations are rejected.
The same cutoff applies to optimization-only and model-assisted annotations,
keeping the acceptance criterion fixed across flywheel rounds.
Human review validates the 2D reference points; this reprojection test
selects the 3D annotations (Section~\ref{sec:coevolve}).

\subsection{Annotation Validation}\label{sec:exp}

We evaluate held-out WinDEX-Hand recordings to assess calibrated RGB
stereo under a consistent validation protocol. Existing headset benchmarks
use monochrome tracking streams or tracker-generated labels
\cite{ref:umetrack,ref:hot3d,ref:egodex}. Independent optical motion capture
quantifies 3D accuracy; human-annotated in-the-wild 2D references
assess image consistency and hand localization. Model comparisons
(Section~\ref{subsec:baseline-comparison}) follow the same protocol.

An independently recorded optical motion-capture set provides 3D reference
measurements (Section~\ref{sec:dataset}).
A rigid target observed by
both systems establishes the transform from the capture frame to the
reference camera frame. This set is excluded from model training and
iterative relabeling.

For motion-capture validation, we report absolute wrist-position error in
the reference camera frame, wrist-relative mean per-joint position error
(MPJPE), and Procrustes-aligned MPJPE (PA-MPJPE), all in millimeters.
MPJPE removes wrist translation; PA-MPJPE aligns rotation, translation and
uniform scale. Compared with our optical motion-capture ground truth,
the retained annotations achieve mean errors of 9.6\,mm for
wrist position, 10.1\,mm for MPJPE and 6.1\,mm for PA-MPJPE across both
hands. These results quantify metric wrist-placement and hand-articulation
accuracy against independent optical ground truth.

For in-the-wild validation, we pool all annotation results, including
those obtained by optimization alone and by model-initialized optimization,
and report the mean 2D reprojection error in pixels. The resulting error
of 7.9 pixels indicates close agreement between the optimized annotations
and image observations across diverse everyday scenes.

\section{WinDEX-Net and the Data--Model Flywheel}\label{sec:flywheel}

WinDEX-Net replaces repeated optimization with learned inference of the
hand trajectory. With calibration $\mathcal{K}$ and the camera trajectory
$\hat{\mathcal{T}}$ estimated by SLAM, it approximates the conditional
MAP solution of (\ref{eq:map}):
\begin{equation}
f_{\phi}(\mathcal{I},\mathcal{K},\hat{\mathcal{T}})
\approx\arg\max_{\mathcal{X}}
\tilde p(\mathcal{X},\hat{\mathcal{T}}\mid\mathcal{I},\mathcal{A}).
\label{eq:learned-map}
\end{equation}
Below, $\mathcal{T}$ denotes this fixed SLAM estimate. The architecture combines
binocular evidence, temporal context and articulated hand geometry;
training applies the corresponding observation, motion and hand-prior
constraints to its predictions. Composing these predictions with SLAM
poses yields world-frame motion. Predictions also initialize further
optimization, and accepted refinements expand training through the
data--model flywheel.

\subsection{WinDEX-Net: The End-to-End Fisheye-Native Model}\label{sec:model}

We observe that standalone 2D hand detectors, without explicit 3D hand
priors, are vulnerable to occlusion and rapid motion and often misestimate
hand-box size. WinDEX-Net therefore integrates localization, pose
estimation and crop refinement within a shared model. It processes native
stereo fisheye images without an external detector, allowing 3D
pose supervision to shape the features used for localization.

\subsubsection{Architecture}

\begin{figure*}[!t]
\centering
\includegraphics[width=\textwidth]{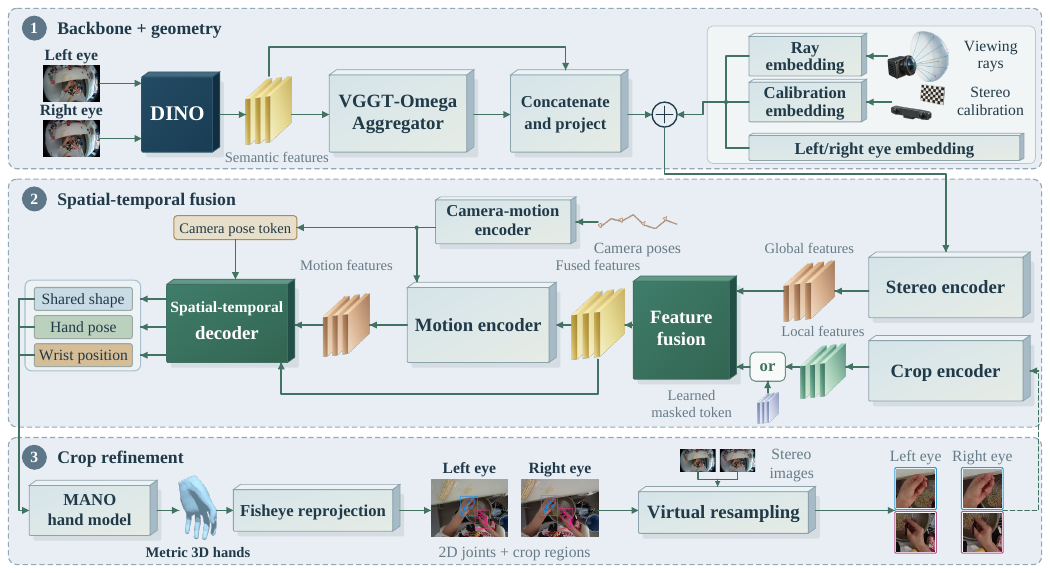}
\caption{WinDEX-Net architecture. Global and local stereo features are fused with temporal motion context to reconstruct metric hand motion. Reprojection closes the iterative crop-refinement loop.}
\label{fig:arch}
\end{figure*}

The architecture comprises geometry-aware visual features, spatial and
temporal fusion, and iterative crop refinement (Fig.~\ref{fig:arch}).
Viewing rays, stereo calibration and eye identity condition the features.
Stereo and local hand encoding feed an odometry-conditioned motion
encoder; joint decoding predicts shared shape $\beta$, articulation
$\theta_t$ and wrist translation $\mathbf{w}_t$. Reprojection updates
the hand crops for subsequent passes.

\begin{table}[!t]
\caption{Depth-branch tap ablation. Derived values in mm.}
\label{tab:tap}
\centering\footnotesize\setlength{\tabcolsep}{4pt}
\begin{tabular}{@{}lccc@{}}
\hline
Variant & \shortstack{Wrist\\error} & MPJPE & PA-MPJPE\\
\hline
Tap layer 4  & 18.00 & 17.08 & 7.92\\
Tap layer 11 (production) & \textbf{15.70} & \textbf{16.68} & \textbf{7.82}\\
Tap layer 17 & 18.70 & 17.78 & 8.12\\
Tap layer 23 (output) & 22.80 & 17.28 & 8.02\\
\hline
\end{tabular}
\end{table}

Geometry-aware visual features. The binocular observation factor
$p_{\mathrm{obs}}$ requires identifying corresponding hand structures
and explaining their positions under calibrated projection. Learned
inference therefore needs semantic cues for correspondence and geometric
cues for metric placement. We combine DINO
appearance features \cite{ref:dinov3} with intermediate features from the
stereo depth branch of VGGT-Omega \cite{ref:vggt}. This choice aims to
balance geometry and semantics: early layers emphasize appearance but
provide limited geometric context, whereas later layers specialize in
depth prediction and may suppress texture and other cues useful for hand
reconstruction. A direct DINO skip connection supplies complementary
semantic information. The relative differences in Table~\ref{tab:tap}
favor the intermediate-layer selection across wrist error, MPJPE and
PA-MPJPE.

To expose the projection geometry of (\ref{eq:stereo}) to learned
inference, we condition image features on per-patch viewing rays and stereo
calibration. Rays account for the nonuniform angular sampling of fisheye
pixels; calibration supplies the baseline and relative camera pose needed
for metric depth. An eye embedding distinguishes the two views. Let $\mathbf{F}^{e}_t$
denote the aggregated features for eye $e$ at time $t$, $\mathbf{r}^{e}$
the ray directions, and
$\mathcal{K}=(\mathtt{K}^l,\mathtt{K}^r,\mathtt{R}_{lr},\mathbf{t}_{lr})$
the calibration. The conditioned features are

\begin{equation}
\tilde{\mathbf{F}}^{e}_t=\mathbf{F}^{e}_t
+\phi_{\mathrm{ray}}(\mathbf{r}^{e})
+\phi_{\mathrm{cal}}(\mathcal{K})
+\mathbf{e}_{e},
\label{eq:cond}
\end{equation}

where $\phi_{\mathrm{ray}}$ and $\phi_{\mathrm{cal}}$ are learned maps and
$\mathbf{e}_e$ is the eye embedding.

Spatial and temporal fusion. The model first establishes stereo
correspondence, enriches joint features with crop detail, and then
aggregates motion across the clip. These operations use residual attention
blocks that combine within-view context with information from the
corresponding token sets:

\begin{equation}
\begin{split}
\mathbf{X}' &= \mathbf{X}+\textstyle\sum_{a\in\{\mathrm{in},\mathrm{cr}\}}
  \mathrm{MHA}_a(\mathrm{LN}\,\mathbf{X}),\\
\mathrm{Blk}(\mathbf{X}) &= \mathbf{X}'+\mathrm{FFN}(\mathrm{LN}\,\mathbf{X}').
\end{split}
\label{eq:blk}
\end{equation}

Stereo encoding. To combine the two observations coupled by
$p_{\mathrm{obs}}$ in (\ref{eq:stereo}), within-eye attention captures hand
structure and cross-eye attention exchanges corresponding evidence.
Joint queries $\mathbf{q}_J$ with hand embeddings $\mathbf{e}_h$ read
this representation into a compact set of joint features:

\begin{align}
\mathbf{G}_t &= \mathrm{Blk}^{\times4}\big(\tilde{\mathbf{F}}^{\,l}_t,
   \tilde{\mathbf{F}}^{\,r}_t\big)\in\mathbb{R}^{2\times P\times d},\\
\mathbf{j}^{\,e}_{t} &= \mathrm{MHA}\big(\mathbf{q}_{J}+\mathbf{e}_{h},\,
   \mathbf{G}^{e}_t,\,\mathbf{G}^{e}_t\big)\in\mathbb{R}^{H\times J\times d}.
\end{align}

Here $P$ and $d$ denote the patch count and feature width, and $H$ and
$J$ the numbers of hands and joints per hand. $\mathbf{G}_t$ contains
stereo patch features, and $\mathbf{j}^{\,e}_t$ contains one feature per
hand and joint in each eye. A dedicated 2D bounding-box head branches from
the stereo image encoder, pooling joint-query features per hand to predict
four normalized box coordinates in each fisheye view. Auxiliary 2D
keypoints $\hat{\mathbf{u}}^{e}_t$ further supervise localization. Wrist tokens also provide
a stereo wrist estimate. Auxiliary 2D supervision and decoded-hand
reprojection enforce the observation consistency in (\ref{eq:stereo});
triangulation additionally provides a metric position estimate:

\begin{equation}
\hat{\mathbf{P}}^{\mathrm{tri}}_t=\mathrm{Tri}\Big(\pi^{-1}_{\mathrm{KB},l}\big(\hat{\mathbf{u}}^{l}_t\big),\,
\pi^{-1}_{\mathrm{KB},r}\big(\hat{\mathbf{u}}^{r}_t\big);\,\mathtt{R}_{lr},\mathbf{t}_{lr}\Big).
\end{equation}

Continuous localization coordinates are unprojected to calibrated camera
rays. Differentiable stereo triangulation passes these estimates to the
decoder without detachment, so 3D and reprojection losses train localization
and reconstruction jointly through MANO decoding.

\begin{figure}[!t]
\centering
\includegraphics[width=\columnwidth]{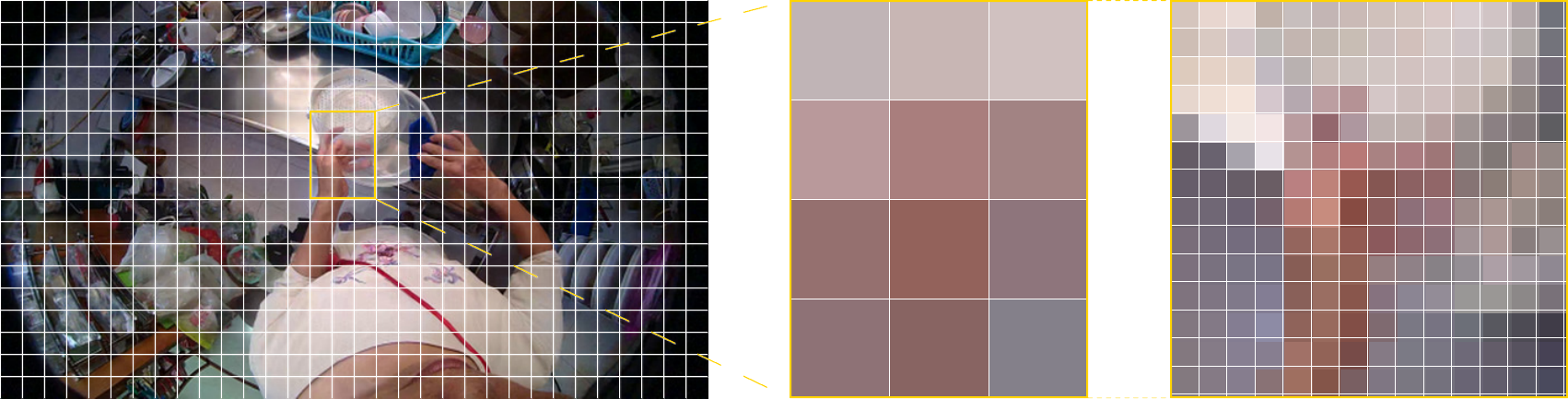}
\caption{Hand token resolution. Left: a $512\times288$ view encoded as $32\times18=576$ patch tokens; the highlighted hand region spans 12 tokens. Middle: the same region shown by mean patch color. Right: resampling to a $224\times224$ hand crop yields $14\times14=196$ tokens, approximately sixteen times as many for this region.}
\label{fig:tokens}
\end{figure}

\begin{figure}[!t]
\centering
\includegraphics[width=\columnwidth]{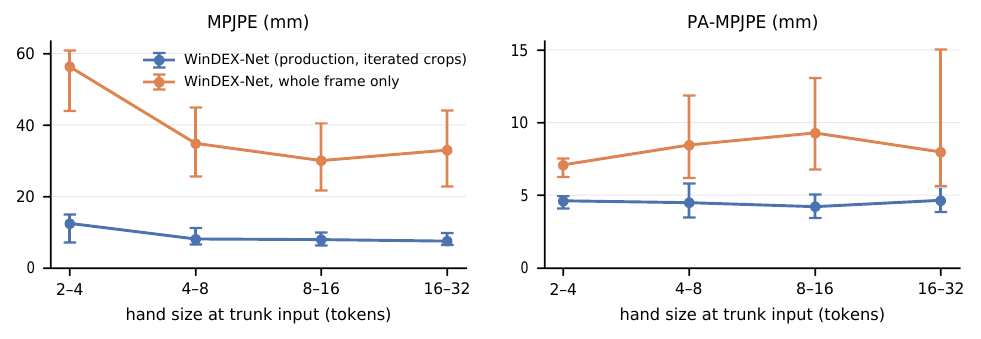}
\caption{Effect of crop refinement. Median and interquartile ranges of MPJPE and PA-MPJPE for WinDEX-Net with and without the crop branch.}
\label{fig:tokcurve}
\end{figure}

Local hand encoding. A fixed token grid limits the finger detail available
to $p_{\mathrm{obs}}$ for small hands. With $16\times16$-pixel patches,
a $512\times288$ full-frame view yields $32\times18=576$ tokens across the
scene. The hand in Fig.~\ref{fig:tokens} occupies only 12 tokens; resampling
to a $224\times224$ crop yields $14\times14=196$ dedicated tokens,
approximately sixteen times as many. Higher errors without crops,
especially for small hands (Fig.~\ref{fig:tokcurve}), motivate the local
branch. We resample virtual-pinhole crops from the original images and
encode them with a frozen DINO encoder. Joint queries read the local
detail, and cross-hand attention provides context for overlapping crops.
A softmax expectation over heatmap cells and a bounded sub-cell
offset yield continuous 2D coordinates, preserving localization gradients.
The local read-out $\mathbf{c}^{\,e}_t$ is fused with
the stereo joint features through a learned projection,
$\mathbf{f}^{\,e}_t=\mathtt{W}[\mathbf{j}^{\,e}_t\,\Vert\,\mathbf{c}^{\,e}_t]$,
so temporal inference receives both scene context and detailed articulation.

Motion encoding. The factor
$p_{\mathrm{dyn}}(\mathbf{x}_t\mid\mathbf{x}_{t-1},\mathbf{x}_{t-2})$
couples successive hand states. The motion encoder combines preceding and
subsequent observations to interpret ambiguous or occluded frames, with temporal
regularization constraining the decoded trajectory. Dense joint--frame
attention is costly, so we pool joint features into one register per hand
and frame before exchanging information across time and hands.
The camera-pose factor $p(\mathcal{T}\mid\mathcal{I},\mathcal{A})$
is estimated by SLAM. Conditioning registers on $\mathcal{T}$ helps separate
headset motion from hand motion. Each joint then attends to its hand's
registers to recover temporal context.
With pooling query $\mathbf{p}_h$, time and hand embeddings $\mathbf{e}_T$
and $\mathbf{e}_h$, and a trajectory encoder $\psi$, this gives

\begin{align}
\mathbf{r}_{t,h} &= \mathrm{MHA}\big(\mathbf{p}_h,\,\mathbf{f}_{t,h},\,
   \mathbf{f}_{t,h}\big),\\
\mathbf{R} &= \mathrm{Blk}^{\times4}\big(\mathbf{r}+\mathbf{e}_{T}+\mathbf{e}_{h}
   +\psi(\mathcal{T})\big)\in\mathbb{R}^{T\times H\times d},\\
\mathbf{m}_{t,h} &= \mathbf{f}_{t,h}+\mathrm{MHA}\big(\mathrm{LN}\,\mathbf{f}_{t,h},\,
   \mathbf{R}_{1:T,h},\,\mathbf{R}_{1:T,h}\big).
\end{align}

Here $\mathbf{R}$ contains the mixed registers and $\mathbf{m}_{t,h}$
the temporally updated joint features.

Joint decoding. The product of factors in (\ref{eq:map}) constrains a
common hand state. We therefore fuse stereo and temporal features with a
camera token before decoding shape, pose and wrist position together.
Alternating spatial and temporal attention captures within-frame hand
structure and cross-frame motion without a dense joint--time attention
matrix. This attention decomposition replaces an $O((T(2J{+}1))^2)$ interaction with
spatial and temporal interactions of sizes $O((2J{+}1)^2)$ and $O(T^2)$.
MANO supplies the articulated geometric parameterization. The prior
$p_{\mathrm{pose}}(\theta_t)$ acts through pose regularization during
training, favoring plausible articulation when image evidence is weak.
Shape regularization represents $p_\beta$. A single $\beta$ is shared
across the clip as in (\ref{eq:state}), preventing changing hand shape from
substituting for motion. Pose and wrist translation vary by frame.
Queries $\mathbf{q}_\beta$, $\mathbf{q}_\theta$ and
$\mathbf{q}_{\mathbf{w}}$ decode these quantities from the fused tokens:

\begin{align}
\mathbf{s}_t &= \mathrm{MLP}\big([\mathbf{f}^{\,l}_t\Vert\mathbf{f}^{\,r}_t\Vert
   \mathbf{m}^{\,l}_t\Vert\mathbf{m}^{\,r}_t]\big)+\mathbf{e}_{J},\\
\mathbf{S} &= \big(\mathrm{Tf}_{\mathrm{time}}\circ\mathrm{Tf}_{\mathrm{sp}}\big)^{\times2}
   \big([\mathbf{s}_{1:T}\,\Vert\,\psi_{\mathrm{cam}}(\mathcal{T})]\big),\\
(\beta,\theta_t,\mathbf{w}_t) &= \mathcal{D}\big(\mathbf{q}_\beta,\mathbf{q}_\theta,
   \mathbf{q}_{\mathbf{w}};\,\mathbf{S}\big).
\label{eq:decode}
\end{align}

Iterative crop refinement. The current hand estimate determines where to
recover additional evidence for $p_{\mathrm{obs}}$. The first pass uses
full-frame features and a learned masked token. Its decoded MANO hand is
reprojected into both fisheye views to define virtual-pinhole crops, which
are resampled directly from the original images. Subsequent passes encode
these crops and repeat (\ref{eq:decode}), updating crop placement from the
refined pose. Localization and reconstruction are thus refined within
the same network. Cached trunk features limit the cost of refinement;
when the window advances, only features for the new observations are
computed.

Fig.~\ref{fig:tokcurve} evaluates the effect of local crop refinement
under matched fine-tuning. The crop branch reduces
MPJPE and PA-MPJPE across the tested hand-size groups. Its largest MPJPE
reduction occurs for hands occupying few trunk tokens. The results support
local resampling for articulation recovery, including hands that are
poorly resolved in the full-frame representation.

\subsubsection{Training}

Both passes share weights and receive intermediate and decoded supervision.
Label-based crops are perturbed in location and supplied distance to mimic
first-pass errors and discourage reliance on crop scale as a depth cue.
Withholding crops and substituting the learned masked token supports
initial inference and repeated refinement.
The crop DINO backbone is frozen; its read-out, localization heads and
fusion layers train jointly with the decoder. Crop-placement updates are
not unrolled for back-propagation.
The joint objective combines metric supervision with image, parameter,
stereo and temporal constraints:

\begin{equation}
\mathcal{L} = \lambda_{\mathrm{3D}}\mathcal{L}_{\mathrm{3D}}
 + \lambda_{\mathrm{2D}}\mathcal{L}_{\mathrm{2D}}
 + \lambda_{\mathrm{par}}\mathcal{L}_{\mathrm{par}}
 + \lambda_{\mathrm{st}}\mathcal{L}_{\mathrm{st}}
 + \lambda_{\mathrm{tm}}\mathcal{L}_{\mathrm{tm}},
\label{eq:loss}
\end{equation}

$\mathcal{L}_{\mathrm{3D}}$ transfers the optimized metric geometry into
the network. The remaining losses carry the factors of (\ref{eq:map}) into
joint training: $\mathcal{L}_{\mathrm{2D}}$ and $\mathcal{L}_{\mathrm{st}}$
enforce the binocular agreement described by $p_{\mathrm{obs}}$;
$\mathcal{L}_{\mathrm{par}}$ imposes the articulation and shape constraints
of $p_{\mathrm{pose}}$ and $p_\beta$; and $\mathcal{L}_{\mathrm{tm}}$
penalizes temporal variation associated with $p_{\mathrm{dyn}}$ while
preserving rapid motion. The negative-log interpretation of the product
in (\ref{eq:map}) motivates their weighted sum: observations, hand priors
and motion constrain the same decoded trajectory together. Auxiliary masked reconstruction
trains the motion registers, and a separately trained confidence head
estimates reconstruction reliability.

Appearance and image-space augmentations cover illumination changes,
sensor noise, compression, gloves, motion blur and occlusion. We preserve
the calibrated fisheye geometry and omit in-plane roll augmentation of
the virtual crops.

\subsection{Iterative Annotation and Independent Audit}\label{sec:coevolve}\label{subsec:audit}

Starting with 20 hours of recordings annotated by optimization,
Algorithm~\ref{alg:flywheel} expands the labeled corpus.
WinDEX-Net predicts initializations for unlabeled recordings.
Reviewers first select reliable, independently detected 2D keypoints.
The pipeline of Section~\ref{sec:annotate} refines the initial 3D estimates
against these references and filters the results by reprojection error.
The accepted image--3D-label pairs enter the next training round.

\begin{algorithm}[!t]
\caption{Iterative model-assisted annotation}
\label{alg:flywheel}
\begin{algorithmic}[1]
\STATE \textbf{input:} dataset $\mathcal{C}$, initial subset $\mathcal{S}_0 \subset \mathcal{C}$,
       audit set $\mathcal{A}$ (never trained on)
\STATE $\mathcal{U} \leftarrow \textsc{HumanScreen2D}(\textsc{Detect2D}(\mathcal{C}))$
\STATE $\mathcal{L}_0 \leftarrow \textsc{Annotate3D}(\mathcal{S}_0,\mathcal{U})$
       \hfill $\triangleright$ fit and filter 3D against $\mathcal{U}$
\FOR{$k = 0,1,2,\dots$}
  \STATE $f_k \leftarrow \textsc{Train}(\mathcal{L}_k)$
         \hfill $\triangleright$ amortize the solver
  \STATE $e_k \leftarrow \textsc{Audit}(f_k, \mathcal{A})$
  \IF{$e_k$ has stopped improving}
    \STATE \textbf{return} $f_k, \mathcal{L}_k$
  \ENDIF
  \STATE $\tilde{\mathcal{L}} \leftarrow \{f_k(c) : c \in \mathcal{C}\}$
         \hfill $\triangleright$ model initialization per clip
  \STATE $\mathcal{L}_{k+1} \leftarrow \textsc{Refine}(\tilde{\mathcal{L}},\mathcal{U})$
         \hfill $\triangleright$ re-optimize under stereo geometry,
  \STATE \hfill independent 2D keypoints and the frozen pose prior
  \STATE $\mathcal{L}_{k+1} \leftarrow \textsc{ReprojectionFilter3D}(\mathcal{L}_{k+1},\mathcal{U})$
\ENDFOR
\end{algorithmic}
\end{algorithm}

Each refinement round re-evaluates the factors of (\ref{eq:map}) on the
model-initialized trajectory: independent 2D keypoints provide
$p_{\mathrm{obs}}$ targets, the frozen pose prior constrains articulation,
and temporal refinement enforces cross-frame consistency. Reprojection
against human-screened keypoints selects the refined 3D labels. Network
fitting and geometric relabeling alternate in this
expectation--maximization-style schedule (Section~\ref{sec:overview}).
An independent
motion-capture set is excluded from fitting and audits each round; iteration
stops when audit error no longer improves. Better initializations can allow
additional clips to pass optimization and filtering, expanding training
coverage. The annotation corpus $\mathcal{C}$ excludes the held-out,
human-annotated in-the-wild test set (Section~\ref{sec:dataset}).

\subsection{The WinDEX-Hand Dataset}\label{sec:dataset}

\begin{figure}[!t]
\centering
\includegraphics[width=\columnwidth]{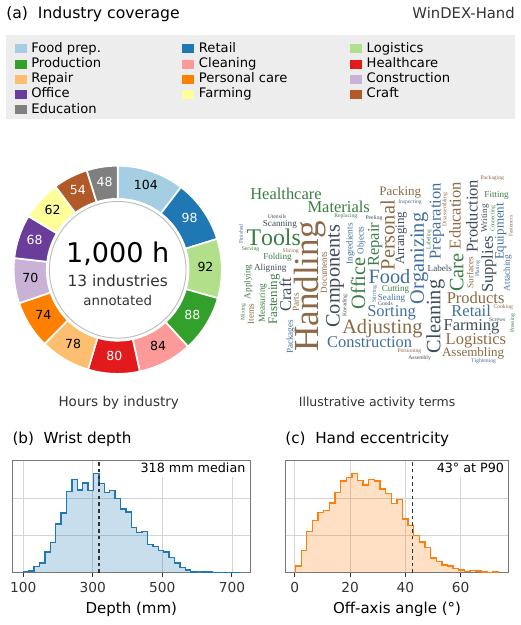}
\caption{WinDEX-Hand training-corpus composition. Statistics exclude the human-annotated test set. (a) Industry hours and an illustrative activity cloud. (b) Wrist depth. (c) Off-axis hand angle.}
\label{fig:datastats}
\end{figure}

WinDEX-Hand separates training supervision from evaluation references
through three data components. The training corpus contains 1{,}000 hours of in-the-wild
pseudo-ground-truth annotations from the data--model loop, collected from
more than 200 operators across more than 1{,}000 scenes and more than
5{,}000 distinct short-range tasks. Drawing on O*NET occupational and
work-activity information \cite{ref:onet}, we organize collection into 13
scene categories: food preparation, retail, logistics, production,
cleaning, healthcare, repair, personal care, construction, office work,
farming, craft and education (Fig.~\ref{fig:datastats}a).

For held-out in-the-wild 2D evaluation, we human-annotate 20 hours collected
separately across the same scene categories. Sapiens2~\cite{ref:sapiens2} and
YOLO11-pose~\cite{ref:yolo11} initialize 2D keypoints. Ordered from low to
high detection confidence, samples undergo human annotation of both
keypoints and bounding boxes, followed by manual review to establish the
test references. The set provides 2D annotations only, with no 3D ground
truth, and is excluded from model training, baseline fine-tuning and the
annotation flywheel.

The optical motion-capture set supplies independent 3D ground truth from
10 volunteers performing 38 task types over 20 hours. We recreate six
everyday settings within the optical capture volume: food preparation,
logistics packing, a retail counter, workshop repair, office desk work
and cleaning. Participants wear CyberCap while optical tracking provides
per-frame 21-keypoint hand positions and validity flags. Activities follow
the same task inventory as the in-the-wild collection. A hardware trigger
synchronizes recordings at each session's start; wrist-velocity
cross-correlation refines temporal alignment offline. The set is excluded
from model training and iterative relabeling. Spatial registration and
accuracy assessment are described in Section~\ref{sec:exp}.

Analysis of the collected in-the-wild data shows a median wrist depth
of 318\,mm, with the central 80\% of hand-frames between 215 and 455\,mm (Fig.~\ref{fig:datastats}b,c), consistent
with the near-field manipulation activities represented in the corpus. Half of
hand-frames are more than 25\degs\ off the optical axis, and one in ten
exceeds 43\degs, underscoring the need to reconstruct hands in peripheral
fisheye regions.

\section{Distillation and Robot Deployment}\label{sec:deliver}

Hand2Policy requires accurate hand trajectories from recorded demonstrations
for policy learning, whereas Hand2Teleop requires low-latency responses to
live human motion. These distinct demands motivate the WinDEX-D family:
WinDEX-Window uses bidirectional, non-causal temporal attention to improve
offline reconstruction accuracy by integrating past and future observations;
WinDEX-Stream uses causal streaming inference to support real-time control
without future frames. Both reconstruct metric hand poses in the reference
camera frame. Combined with visual--inertial camera poses, these estimates
provide world-frame 3D hand representations for retargeting to robot kinematics
(Section~\ref{sec:downstream}).

\subsection{The WinDEX-D Family}\label{subsec:distill}

The two members use different visual backbones, temporal architectures and
distillation targets. WinDEX-Window is a 116M-parameter, 32-frame student that
compresses WinDEX-Net through per-joint feature matching while retaining its
reconstruction stages. WinDEX-Stream is an 8.92M-parameter causal student
trained from a frozen WinDEX-Window checkpoint through feature and motion-output
supervision. The former preserves window-based MANO reconstruction for offline
use; the latter learns recurrent prediction of metric 3D joints for live control.

\subsubsection{WinDEX-Window: Offline Reconstruction}\label{subsec:window-distill}

The main cost of WinDEX-Net lies in the VGGT-Omega trunk's global attention over
both views' patches. To reduce this cost while retaining the learned hand
reconstruction stages, WinDEX-Window replaces the trunk at its per-joint token
interface. A DINOv3 ViT-S/16 \cite{ref:dinov3} encodes each eye, followed by an
image encoder structurally identical to the teacher's for cross-view fusion.
The crop branch, geometry, motion encoder and decoder are retained. The student
inherits the teacher's cross-eye and read-out weights at initialization;
among the retained modules, only the crop branch's input projection is learned
anew.

This interface also makes distillation practical at corpus scale: per-joint
teacher tensors are compact enough to cache, avoiding repeated teacher
inference during frame-wise student training. We compute targets from clean
images and augment only student inputs, encouraging consistent features under
appearance changes. Interface matching is followed by end-to-end fine-tuning
under (\ref{eq:loss}), because crop processing and geometric refinement also
consume features outside the distilled interface.

Joint, hand and eye identity dominate the interface tensors, so a high raw
cosine can arise from an image-independent constant. We therefore use
intermediate-feature distillation \cite{ref:hinton,ref:fitnets} to match centered,
channel-normalized features. For teacher tensor $\mathbf{z}$ and student tensor
$\hat{\mathbf{z}}$, the student maximizes

\begin{equation}
s = \cos\!\left(\frac{\mathbf{z}-\bar{\mathbf{z}}_{ehj}}{\boldsymbol{\sigma}},\
\frac{\hat{\mathbf{z}}-\bar{\mathbf{z}}_{ehj}}{\boldsymbol{\sigma}}\right),
\label{eq:distill}
\end{equation}
where $\bar{\mathbf{z}}_{ehj}$ is the teacher mean per eye, hand and joint,
and $\boldsymbol{\sigma}$ is its per-channel standard deviation. Providing
the mean to the student as a fixed buffer separates this component from the
image-dependent residual being matched. Channel standard deviations span more
than an order of magnitude; scaling by $\boldsymbol{\sigma}$ reduces the
dominance of high-variance channels in the cosine and its gradients.

\subsubsection{WinDEX-Stream: Causal Streaming}\label{subsec:stream-distill}

\begin{figure}[!t]
\centering
\includegraphics[width=\columnwidth]{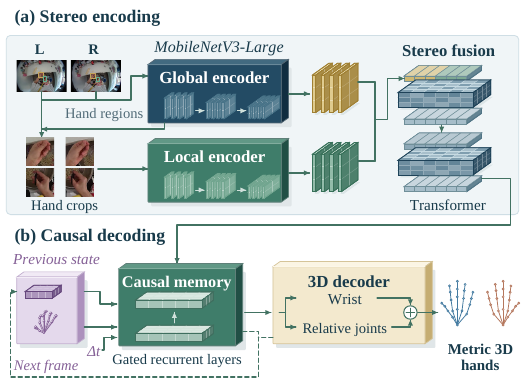}
\caption{WinDEX-Stream architecture. Global and hand-crop features are fused across stereo views. Recurrent state and pose feedback support causal prediction of 21 metric 3D joints per hand.}
\label{fig:teleop_streaming}
\end{figure}

To provide temporal context without future frames, WinDEX-Stream processes
each stereo pair once and carries history in recurrent state. Separate
MobileNetV3-Large encoders~\cite{ref:mobilenetv3} for full images and hand crops combine image context
with local hand detail; a two-layer stereo Transformer fuses the views,
and a two-layer causal GRU supplies temporal context to a direct decoder for
21 metric 3D joints per hand. Predicted regions guide the crops, and recurrent
state and the previous pose pass to the next update, supporting per-frame
causal inference. Fig.~\ref{fig:teleop_streaming} shows the architecture;
Section~\ref{subsec:teleop} demonstrates its use for live robot control.

Whereas WinDEX-Window inherits the teacher's reconstruction modules and first
matches their feature interface, WinDEX-Stream learns a separate joint decoder
using cached outputs of the frozen WinDEX-Window teacher. Its distillation
loss combines absolute wrist-position and wrist-relative joint errors with
feature and wrist-velocity matching. A learned projection maps student features
to the teacher interface, where squared error is measured against teacher
features standardized using fixed training-set statistics. Wrist and relative
joint targets use $L_1$ losses; wrist velocities use a smooth-$L_1$ loss.
This transfers metric placement, articulation and motion alongside visual
features, without requiring the causal student to reproduce MANO parameters.

The distillation terms are optimized jointly with coarse and crop-level 2D
heatmap, coordinate, reprojection and visibility supervision from the screened
annotations. Training processes each 32-frame clip in chronological order,
carrying state across four 8-frame segments and truncating gradients at segment
boundaries. The offline teacher can supply targets using temporal context,
while the student receives only the current stereo pair and its retained past
state. Thus, Window uses feature matching followed by reconstruction
fine-tuning; Stream jointly learns features and causal motion outputs.

\subsection{Hand2Policy and Hand2Teleop}\label{subsec:exec}

Retargeting separates wrist-based arm targets from wrist-relative hand
articulation to map the shared motion representation to each robot's
kinematics. In Hand2Teleop, a local server runs WinDEX-Stream alongside
real-time VIO. WinDEX-Stream reconstructs metric 3D keypoints, and VIO
camera poses transform them into a global coordinate frame to reduce
jitter caused by head motion. The controller uses global wrist motion
relative to a session anchor for arm targets and wrist-relative
articulation for hand targets. Filtering, transient jump rejection and
retention of the last accepted target reduce the influence of unreliable
observations on commands.

In Hand2Policy, WinDEX-Window reconstructs recorded trajectories
for reinforcement-learning transfer. Retargeting each clip once produces a
robot-specific reference that a tracking policy can reuse for replay, reset
or training.

Both interfaces obtain arm targets from wrist displacement through damped
least-squares inverse kinematics. Hand targets use analytical retargeting of
the dex-retargeting family \cite{ref:handmap,ref:anyteleop}. The live Wuji
hand uses the vendor implementation \cite{ref:wujiretarget}, and the
simulated Inspire and Wuji hands use the trajectory imitator's retargeting
stage \cite{ref:maniptrans} (Section~\ref{subsec:rl}).

\section{Experiments}\label{sec:downstream}

\begin{figure*}[!p]
\centering
\includegraphics[width=\textwidth]{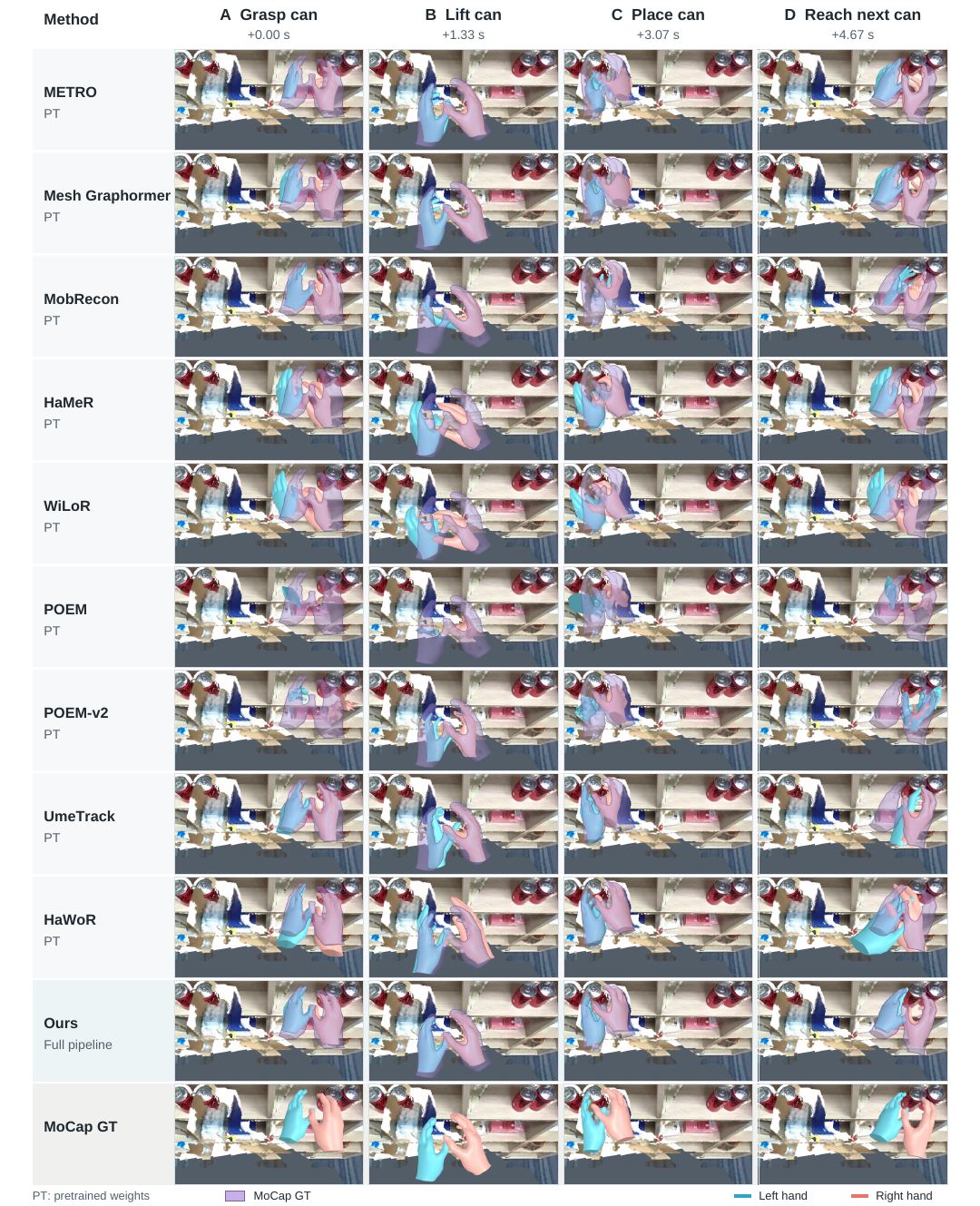}
\caption{Comparison against motion-capture ground truth. Columns show grasping, lifting, placing and reaching during can sorting.}
\label{fig:mocap_comparison}
\end{figure*}

\begin{figure*}[!p]
\centering
\includegraphics[width=\textwidth]{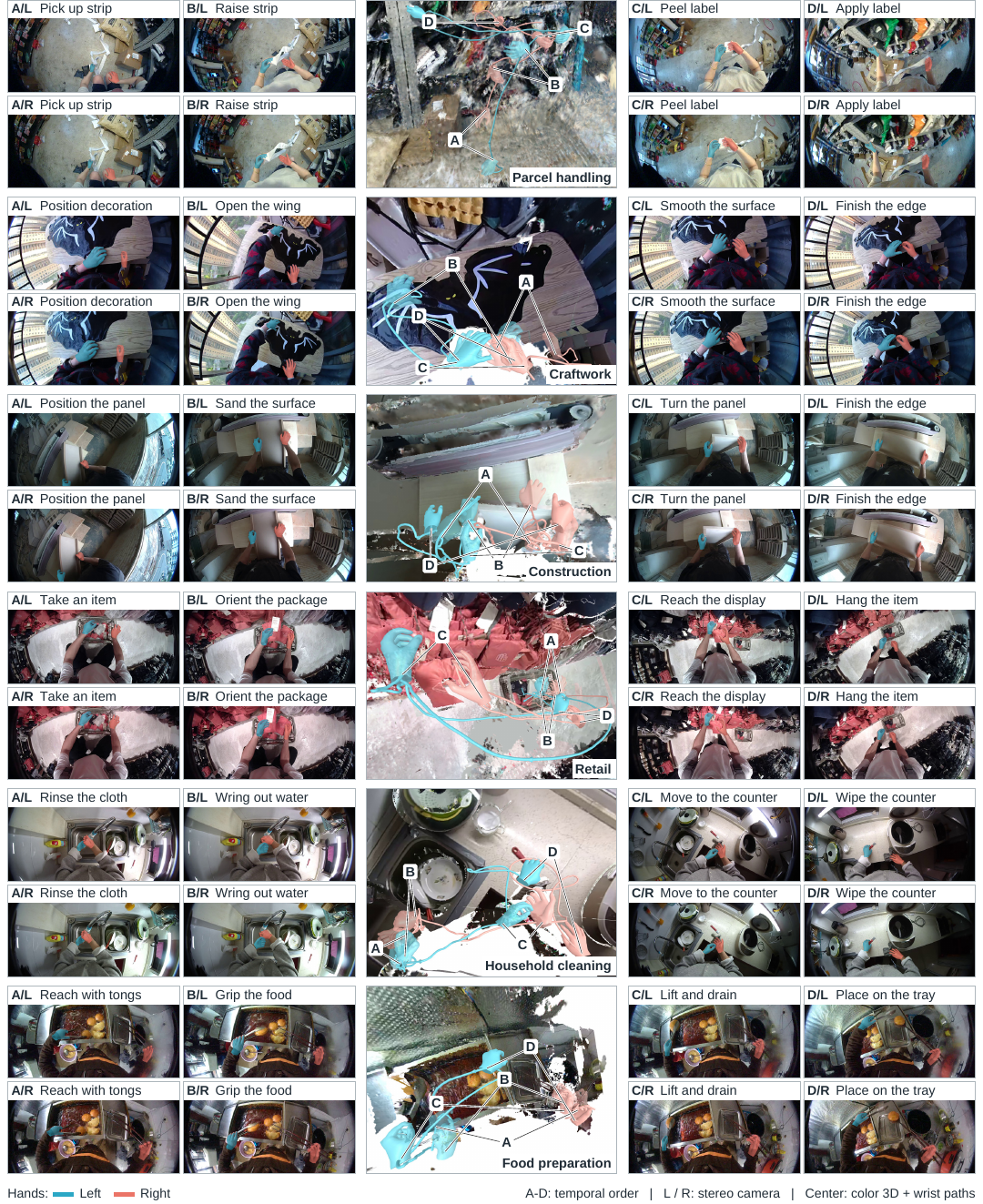}
\caption{In-the-wild reconstruction. Each row pairs stereo observations A--D with world-frame hands and wrist paths. L/R denote cameras; cyan/coral denote left/right hands.}
\label{fig:wild_pc}
\end{figure*}

\begin{table}[!t]
\caption{Speed--accuracy comparison. WinDEX-Net and WinDEX-Window are evaluated at different crop iterations; WinDEX-Stream uses causal per-frame inference. Throughput is measured in FPS; position errors are in mm.}
\label{tab:tradeoff}
\centering\footnotesize\setlength{\tabcolsep}{2pt}
\begin{tabular}{lrrrr}
\hline
Model, iterations & FPS & \shortstack{Wrist\\error} & MPJPE & PA-MPJPE\\
\hline
WinDEX-Net, 2 & 9.50 & 25.9 & 17.8 & 7.8 \\
WinDEX-Net, 3 & 8.17 & 20.4 & 15.5 & 7.8 \\
WinDEX-Net, 4 & 7.38 & 18.9 & 15.4 & 7.8 \\
WinDEX-Window, 2 & 43.61 & 27.1 & 20.1 & 8.1 \\
WinDEX-Window, 3 & 35.94 & 23.8 & 18.7 & 8.1 \\
WinDEX-Window, 4 & 31.77 & 20.3 & 18.7 & 8.1 \\
\hline
WinDEX-Stream & 171.00 & 42.7 & 33.2 & 12.5 \\
\hline
\end{tabular}
\end{table}

\subsection{Implementation and Evaluation Protocol}

WinDEX-Net and WinDEX-Window are implemented
in PyTorch and trained on four RTX 4090 GPUs. WinDEX-Net uses the VGGT-Omega
trunk, (\ref{eq:loss}), AdamW and \texttt{bf16} autocast, with its pretrained
trunk at one tenth of the head learning rate. WinDEX-Window uses DINOv3
ViT-S/16, first matching cached WinDEX-Net features under (\ref{eq:distill})
and then fine-tuning end to end as described in
Section~\ref{subsec:window-distill}. WinDEX-Stream uses MobileNetV3-Large
encoders and causal recurrent fusion, trained with the feature and motion-output
distillation objectives in Section~\ref{subsec:stream-distill}.

For the WinDEX-Net ablations, we evaluate the teacher's depth-feature tap under
matched initialization and fine-tuning, changing only the selected layer.
We assess crop refinement on held-out clips and group results by projected
hand size to test whether local resampling benefits hands that are poorly
resolved in full-frame features.

The offline evaluations in
Sections~\ref{subsec:reference-configuration} and~\ref{subsec:baseline-comparison}
use an NVIDIA RTX 4090 with FP16 inference, FP32 crop geometry, sequence
batch size one, and 32-frame windows at stride 16 with overlap crossfading.
WinDEX-Net uses the same offline inference protocol for the speed--accuracy
comparison. Reconstruction comparisons use WinDEX-Window with two crop
iterations. Reconstruction is evaluated against optical motion-capture ground
truth and human-annotated 2D references from the held-out in-the-wild test set
(Section~\ref{subsec:baseline-comparison}).

WinDEX-Stream updates from each
incoming stereo pair and retained state, without future observations.
Table~\ref{tab:tradeoff} reports its causal per-frame throughput alongside
source-frame throughput for the offline models. Its use for live control is
evaluated separately in Hand2Teleop (Section~\ref{subsec:teleop}).

Hand2Policy
(Section~\ref{subsec:rl}) uses WinDEX-Window trajectories with the simulation
imitator of \cite{ref:maniptrans} and Inspire and Wuji hands. Hand2Teleop
(Section~\ref{subsec:teleop}) uses WinDEX-Stream on a live platform comprising
a 20-joint Wuji five-finger hand, a seven-joint robot arm and an operator-worn
stereo capture device.

\subsection{Reference Configuration for Subsequent Comparisons}\label{subsec:reference-configuration}

\begin{table*}[!t]
\caption{Motion-capture comparison with prior methods. P: pretrained; FT: fine-tuned. Wrist error is absolute wrist-position error; MPJPE is wrist-relative; PA-MPJPE is Procrustes-aligned. Jitter measures acceleration-based temporal variation. UmeTrack uses 18 corresponding joints; HaWoR uses its camera stage.}
\label{tab:compare}
\centering\small
\setlength{\tabcolsep}{9pt}\renewcommand{\arraystretch}{1.08}
\begin{tabular}{llrrrr}
\toprule
Method & Weights & Wrist error $\downarrow$ & MPJPE $\downarrow$ & PA-MPJPE $\downarrow$ & Jitter $\downarrow$ \\
 & & (mm) & (mm) & (mm) & (m/s$^2$) \\
\midrule
METRO \cite{ref:metro} & P & 152.26 & 46.10 & 16.13 & 19.66 \\
 & FT & 48.34 & 34.20 & 10.33 & 13.37 \\
\addlinespace[2pt]
Mesh Graphormer \cite{ref:meshgraphormer} & P & 114.89 & 36.33 & 13.73 & 18.45 \\
 & FT & 57.25 & 28.26 & 10.41 & 12.19 \\
\addlinespace[2pt]
MobRecon \cite{ref:mobrecon} & P & 75.54 & 45.76 & 13.99 & 9.42 \\
 & FT & 53.51 & 36.86 & 11.66 & 7.21 \\
\addlinespace[2pt]
HaMeR \cite{ref:hamer} & P & 62.43 & 36.50 & 10.90 & 9.72 \\
 & FT & 43.29 & 25.29 & 10.09 & 7.35 \\
\addlinespace[2pt]
WiLoR \cite{ref:wilor} & P & 63.25 & 33.65 & 10.34 & 9.01 \\
 & FT & 38.23 & 24.53 & 10.16 & 6.96 \\
\addlinespace[2pt]
POEM \cite{ref:poem} & P & 104.45 & 137.51 & 38.21 & 20.74 \\
 & FT & 32.17 & 37.56 & 18.80 & 9.51 \\
\addlinespace[2pt]
POEM-v2 \cite{ref:poemv2} & P & 46.26 & 34.22 & 13.54 & 10.63 \\
 & FT & 28.77 & 27.02 & 11.13 & 8.30 \\
\addlinespace[2pt]
UmeTrack \cite{ref:umetrack} & P & 42.59 & 36.80 & 11.62 & 10.83 \\
 & FT & 27.45 & 24.09 & 10.82 & 7.82 \\
\addlinespace[2pt]
HaWoR \cite{ref:hawor} & P & 46.05 & 33.78 & 10.61 & 8.68 \\
 & FT & 31.87 & 23.48 & 9.45 & 5.50 \\
\midrule
\textbf{WinDEX-Window (Ours)} & P & \textbf{15.70} & \textbf{16.68} & \textbf{7.82} & \textbf{3.15} \\
\bottomrule
\end{tabular}
\end{table*}

\begin{table}[!t]
\caption{World-frame motion-capture comparison. Position errors are in mm; RTE is a percentage.}
\label{tab:world_compare}
\centering\small
\setlength{\tabcolsep}{4pt}\renewcommand{\arraystretch}{1.08}
\begin{tabular}{@{}lrrr@{}}
\toprule
Method & W-MPJPE & WA-MPJPE & RTE \\
\midrule
HaWoR (P) & 100.92 & 44.19 & 3.10 \\
HaWoR (FT) & 97.65 & 40.91 & 2.83 \\
\midrule
WinDEX-Window + cuVSLAM & \textbf{37.29} & \textbf{17.69} & \textbf{2.00} \\
\bottomrule
\end{tabular}
\end{table}

Table~\ref{tab:tradeoff} compares WinDEX-Net and WinDEX-Window on three
motion-capture recordings at two, three and four crop iterations.
Additional iterations reduce wrist error with diminishing MPJPE gains,
unchanged PA-MPJPE and lower throughput. WinDEX-Window is substantially
faster than WinDEX-Net with comparable PA-MPJPE. We fix WinDEX-Window at
two crop iterations for subsequent reconstruction comparisons, balancing
accuracy and cost. WinDEX-Stream at 171 FPS serves causal live control,
evaluated through Hand2Teleop (Section~\ref{subsec:teleop}).

\subsection{Comparison with Prior Methods}\label{subsec:baseline-comparison}

Evaluation uses independent optical 3D ground truth and human-annotated
in-the-wild 2D references from recordings held out from training
(Section~\ref{sec:exp}). Generated 3D labels supervise training only.
Camera-frame metrics are absolute wrist-position error,
wrist-relative MPJPE and PA-MPJPE in millimeters, and acceleration-based
jitter in m/s$^2$. We also report 2D reprojection error in pixels.

\begin{table}[!t]
\caption{2D hand detection. In-the-wild boxes are human-annotated. Precision (P), recall (R), F1 (\%).}
\label{tab:hand_detection}
\centering\small
\setlength{\tabcolsep}{4pt}\renewcommand{\arraystretch}{1.05}
\begin{tabular}{@{}lrrrrrr@{}}
\toprule
 & \multicolumn{3}{c}{In the wild (20 h)} & \multicolumn{3}{c}{Motion capture} \\
\cmidrule(lr){2-4}\cmidrule(l){5-7}
Method & P & R & F1 & P & R & F1 \\
\midrule
Sapiens2-1B & 82.09 & \textbf{77.43} & 79.69 & 84.60 & 78.42 & 81.39 \\
YOLO11-pose & 70.31 & 76.37 & 73.21 & 76.08 & 77.37 & 76.72 \\
ViTPose-H & 50.00 & 30.41 & 37.82 & 51.34 & 50.53 & 50.93 \\
WinDEX-Net & \textbf{85.07} & 77.09 & \textbf{80.88} & \textbf{88.24} & \textbf{86.84} & \textbf{87.53} \\
\bottomrule
\end{tabular}
\end{table}

Table~\ref{tab:hand_detection} compares WinDEX-Net's dedicated 2D box head
with Sapiens2-1B~\cite{ref:sapiens2},
YOLO11-pose~\cite{ref:yolo11} and ViTPose-H~\cite{ref:vitpose}
using precision, recall and $\mathrm{F1}=2PR/(P+R)$.
WinDEX-Net leads in precision and F1 on both sets and recall on motion
capture; Sapiens2-1B has slightly higher recall in the wild.

Table~\ref{tab:compare} compares nine reconstruction baselines with
pretrained (P) and fine-tuned (FT) weights, using the same input adapter
and test recordings for both settings. Fine-tuning
uses only the training corpus. Single-image methods use the calibrated left
view, multi-view methods use stereo, and temporal methods receive their
required context. UmeTrack uses 18 corresponding joints; the others use 21.
HaWoR contributes its camera-space estimator to Table~\ref{tab:compare}
and its complete pipeline to the world-frame comparison.

Table~\ref{tab:compare} evaluates camera-frame reconstruction against
optical ground truth; Fig.~\ref{fig:mocap_comparison} shows qualitative
comparisons. WinDEX-Window achieves the lowest wrist error, MPJPE,
PA-MPJPE and jitter, outperforming
all nine baselines in both pretrained and fine-tuned settings. These
results establish state-of-the-art performance across the reported
camera-frame metrics on this benchmark.

\begin{figure}[!t]
\centering
\includegraphics[width=\columnwidth]{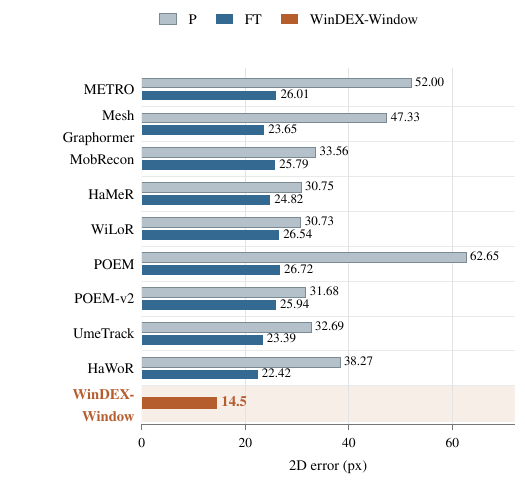}
\caption{Reprojection error on held-out human-annotated in-the-wild keypoints (px; lower is better). P/FT: pretrained/fine-tuned; WinDEX-Window has P only. UmeTrack uses 18 joints; HaWoR uses its camera-space stage.}
\label{fig:external2d}
\end{figure}

We also evaluate world-frame trajectories using world MPJPE (W-MPJPE),
world-aligned MPJPE (WA-MPJPE) and root translation error (RTE), following
HaWoR~\cite{ref:hawor}. W-MPJPE and WA-MPJPE fit a similarity transform
to the first two frames and all frames of each 100-frame segment,
respectively. RTE is the mean root-trajectory error after rigid alignment
without scaling, normalized by cumulative ground-truth path length.
For world-frame evaluation, we use HaWoR's complete hand-reconstruction
and camera-trajectory pipeline. Table~\ref{tab:world_compare} compares it
with WinDEX-Window
composed with cuVSLAM camera poses, which reduces W-MPJPE, WA-MPJPE and RTE by
61.8\%, 56.8\% and 29.3\% relative to fine-tuned HaWoR.

WinDEX-Window's 14.5\,px reprojection error against held-out human-annotated
in-the-wild keypoints, versus 22.42--26.72\,px for fine-tuned baselines
(Fig.~\ref{fig:external2d}), indicates better cross-scene generalization and
image consistency. Fig.~\ref{fig:wild_pc} pairs stereo observations with
world-frame hands and wrist trajectories across parcel handling,
craftwork, construction, retail, cleaning and food preparation.

\subsection{Hand2Policy}\label{subsec:rl}

\begin{figure}[!ht]
\centering
\includegraphics[width=\columnwidth]{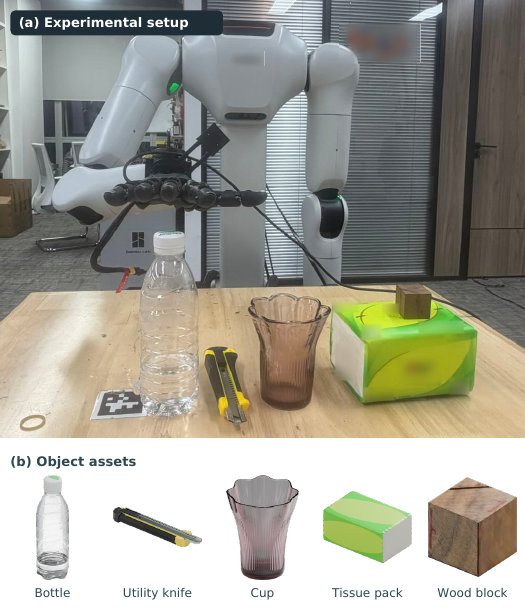}
\caption{Experimental setup. (a) Physical setup before manipulation. (b) Bottle, utility knife, cup, tissue pack and wood block assets.}
\label{fig:rl_initial_workspace}
\end{figure}

Hand2Policy uses reconstructed human motion to learn robot policies.
We study reinforcement learning (RL) as one implementation, using the
ManipTrans trajectory imitator~\cite{ref:maniptrans} driven by retargeted
hand-motion references.
We evaluate trajectory following and grasp--lift--place trials in simulation
and on hardware. Fig.~\ref{fig:rl_initial_workspace} shows the experimental
setup and the five object assets. We create the object assets using
GPT6-astra and Blender, and set their physical parameters from measurements
of the corresponding real objects.

\begin{figure*}[!t]
\centering
\includegraphics[width=\textwidth]{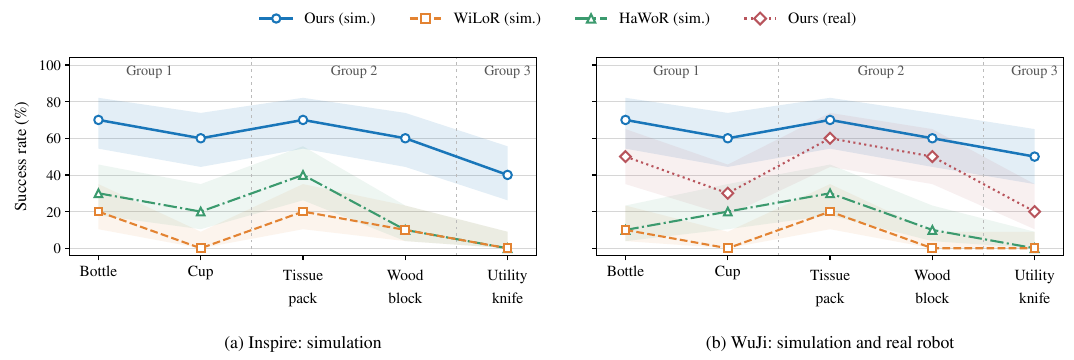}
\caption{Hand2Policy: grasp--lift--place success. Each point reports 10 trials with
object positions randomized within $20\times20$\,cm; success requires grasping,
a 10\,cm lift and placement. Shading shows pointwise 68\% Wilson binomial
confidence intervals. Dashed lines separate the three object groups.
Compared guidance methods share training settings within each hand embodiment:
2{,}000 iterations for Wuji and 5{,}000 for Inspire.
Real-robot trials use Wuji with privileged Gemini 2L/FoundationPose~\cite{ref:foundationpose} localization.}
\label{fig:object_success}
\end{figure*}

All compared guidance methods use identical training settings within each
hand embodiment, differing only in hand-motion references: 2{,}000 iterations
for the electrically driven Wuji hand and 5{,}000 for the tendon-driven
Inspire hand.

Each bottle, cup, tissue pack, wood block and utility knife configuration
has 10 trials per method and hand, with positions randomized within
$20\times20$\,cm. Success requires grasping, a 10\,cm lift and placement;
simulation and hardware use the same protocol (Fig.~\ref{fig:object_success}).

Under these matched training settings, our
hand-motion references achieve the highest observed success rates across
all evaluated objects with both Inspire and Wuji. Averaged across the five
objects and two hands, simulated success reaches 61\%, compared with 8\%
for WiLoR and 17\% for HaWoR (Fig.~\ref{fig:object_success}).

Physical Wuji experiments use an external Gemini 2L camera and
FoundationPose~\cite{ref:foundationpose} for privileged object and robot-arm
localization, approximating the pose information available in simulation.
Transferred policies complete grasp--lift--place trials across all five
objects (Fig.~\ref{fig:object_success}).

\subsection{Hand2Teleop}\label{subsec:teleop}

Unlike DexPilot, which uses a four-camera external RGB-D
rig~\cite{ref:dexpilot}, and BiDex, which uses motion-capture gloves and
teacher arms~\cite{ref:bidex}, our system tracks the operator using only
CyberCap's stereo cameras and onboard IMU. Hand2Teleop maps reconstructed
wrist motion and finger articulation to coordinated arm--hand commands
for live teleoperation.

On an NVIDIA RTX 4090 server, WinDEX-Stream (Section~\ref{subsec:distill}
and Fig.~\ref{fig:teleop_streaming}) reconstructs hands from CyberCap
streams alongside real-time VIO camera-pose estimation. VIO poses place
the hands in a world frame to reduce head-motion jitter. Global wrist
motion drives robot-arm targets through inverse kinematics, and the Wuji
SDK maps finger articulation to hand-joint commands. The arm and hand
receive coordinated commands at 30\,Hz, matching CyberCap's stereo capture
rate, with an approximately 33.3\,ms update period.

\begin{figure}[!ht]
\centering
\includegraphics[width=\columnwidth]{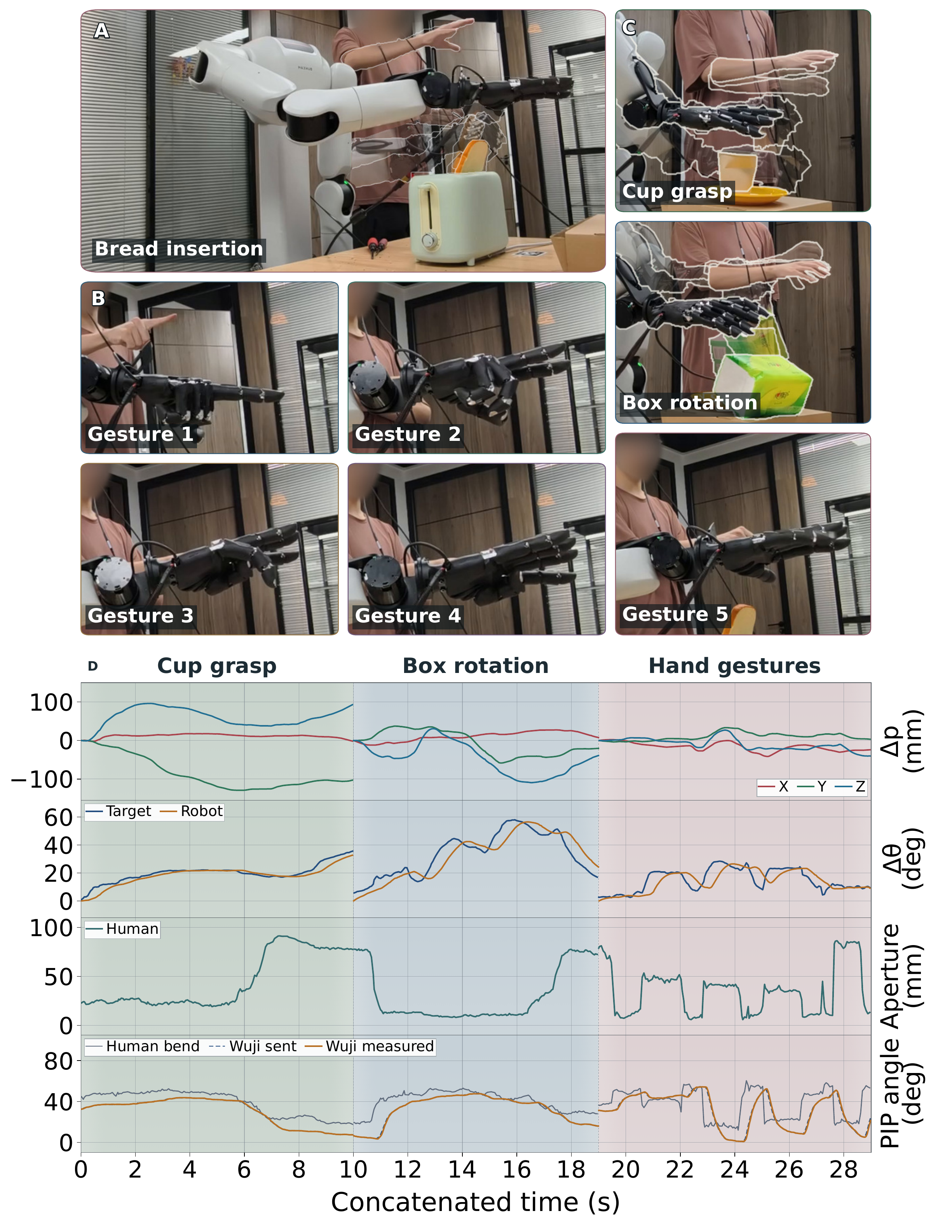}
\caption{Hand2Teleop: live teleoperation. (A) Bread insertion; (B) hand gestures; (C) cup grasp, box reorientation and a hand gesture. (D) Recorded arm and finger motion during cup, box and gesture excerpts; colored bands distinguish separate excerpts.}
\label{fig:teleop_motion}
\end{figure}

Paper-cup grasp and release assess opening and
closing accuracy; tissue-box reorientation and bread insertion assess
wrist-driven arm positioning and orientation; and gestures assess
finger-motion control (Fig.~\ref{fig:teleop_motion}). The corresponding
curves show end-effector displacement and rotation, human hand aperture
and finger bending, and robot joint commands and measurements across
cup, box and gesture excerpts.

\section{Discussion and Limitations}\label{sec:limits}

We identify three directions for future improvement.
First, prolonged occlusion removes visual constraints on
articulation and position. Stronger temporal reasoning and scene-interaction
context are needed for persistent tracking and recovery.

Second, training on CyberCap recordings limits device diversity. Data
spanning different optics, stereo geometries and mounting configurations
are needed to improve and assess cross-device generalization.

Third, combining object geometry, pose and intended interaction with
wrist motion and articulation could improve teleoperation precision
through object-aware motion mapping.

\section{Conclusion}

WinDEX connects scalable in-the-wild egocentric data to metric,
world-frame 3D hand motion and physical robot control and learning.
Large-scale data collection, geometric--temporal annotation and the
WinDEX-Net data--model flywheel form a capture-to-reconstruction pipeline.
WinDEX-Hand combines in-the-wild training annotations with held-out
human-annotated 2D references and independent optical motion-capture ground truth.
WinDEX-D provides fast offline inference with WinDEX-Window and causal
streaming inference with WinDEX-Stream. Reconstruction, Hand2Policy and
Hand2Teleop evaluations demonstrate the recovered motion's utility for
perception and robotic manipulation.

Future work will address long-horizon drift and occlusion, broaden
evaluation across objects, tasks and robot embodiments, and incorporate
human contact and robot tactile data to combine coarse visual positioning
with fine tactile control.


\clearpage
\bibliographystyle{IEEEtran}
\bibliography{refs}

\end{document}